\documentclass[a4paper,parskip=half]{article} 

\usepackage{authblk}

\usepackage[utf8]{inputenc}
\usepackage[T1]{fontenc}
\usepackage{amsmath,mathtools,amssymb,listingsutf8,xspace}
\usepackage{geometry,hyperref,cleveref,xcolor}
\usepackage[textsize=scriptsize
           ,disable
           ]{todonotes}

\usepackage{tikz-cd}
\usepackage{booktabs}

\usetikzlibrary{positioning}

\newcommand*\cmdstyle\texttt
\newcommand*\file\cmdstyle
\newcommand*\literalColor{blue}
\newcommand*\cmd[1]{\cmdstyle{\textcolor{red!85!black}{#1}}}
\newcommand*\cmdline[1]{\cmdstyle{\textcolor{green!70!black}\$ }\cmd{#1}}
\newcommand*\literal[1]{\textcolor{\literalColor}{\cmdstyle{#1}}}
\newcommand*\api[1]{\textcolor{purple}{\cmdstyle{#1}}}
\newenvironment{cmdhelp}{\begin{quote}\footnotesize}{\end{quote}}

\newcommand\removed[1]{}
\newcommand\added[1]{#1}
\newcommand\changedto[2]{%
    \removed{#1}%
    \added{#2}}
\newcommand\changed[1]{{\color{blue}#1}}

\newcommand*\eqdef=

\newcommand*\query{\operatorname{\mathsf{query}}}
\newcommand*\assert{\operatorname{\mathsf{assert}}}
\newcommand*\formula{\operatorname{\varphi_{\mathit{post}}}}

\newcommand*\regmax[2]{[[{#1}]]_{#2}}
\newcommand*\objv{o}
\newcommand*\option[1]{\operatorname{\mathsf{#1}}} 
\newcommand*\optionval[2]{\operatorname{\mathsf{#1}\,\,\mathsf{#2}}} 
\newcommand*\suffix[1]{\operatorname{\mathsf{#1}}} 
\newcommand*\mode[1]{\operatorname{\mathsf{#1}}} 
\newcommand*\package[1]{\operatorname{\mathsf{#1}}} 
\newcommand*\module[1]{\operatorname{\mathsf{#1}}} 
\newcommand*\dtype[1]{\operatorname{\mathsf{#1}}} 
\newcommand*\precision[1]{\operatorname{\mathsf{#1}}} 
\newcommand*\model[1]{\operatorname{\mathsf{#1}}} 
\newcommand*\speckey[1]{\operatorname{\mathsf{"{#1}"}}} 
\newcommand*\specval[1]{\operatorname{\mathsf{{#1}}}} 
\newcommand*\operator[1]{\operatorname{\mathsf{{#1}}}} 

\newcommand*\Solver{Symbolic Machine Learning Prover\xspace}
\newcommand*\SolverAbbrvText{SMLP}
\newcommand*\SolverAbbrv{\SolverAbbrvText\xspace}
\newcommand*\SolverVersion{v0.1}

\newcommand*\progmrc{smlp-mrc.sh}
\newcommand*\provenn{prove-nn.py}
\newcommand*\trainnn{train-nn.py}

\newcommand*\todozk[2][]{\todo[color=yellow!30,tickmarkheight=.2em,size=\scriptsize,#1]{ZK: #2}}
\newcommand*\todokk[2][]{\todo[color=purple!20,tickmarkheight=.2em,size=\scriptsize,#1]{KK: #2}}

\newtheorem{thm}{Theorem} 

\newtheorem{defn}[thm]{Definition}
\newtheorem{exmpl}[thm]{Example}

\newcommand*\KK{\todokk}
\newcommand*\ZK{\todozk}

\newcommand{\delete}[1]{}

\title{SMLP: \Solver  \\
(User Manual)
}

\author[1]{Franz Brau\ss{}e}
\author[2]{Zurab Khasidashvili}
\author[3]{Konstantin Korovin}
\affil[1,3]{The University of Manchester, UK}
\affil[2]{Intel, Israel}

\begin{document}
\maketitle
\begin{abstract}

\delete{
\Solver (\SolverAbbrv) is a collection of tools for reasoning about machine
learning models. \SolverAbbrv is based on SMT solver(s). In this document we
describe functionality for computing safe and stable regions of neural network
models satisfying optput specifications. This corresponds to solving
$\epsilon$-guarded $\exists^*\forall^*$ formulas over NN
representations~\cite{DBLP:conf/fmcad/BrausseKK20}.
}

\SolverAbbrv: \Solver is an open source tool for exploration and optimization of systems represented by machine learning models.\footnote{SMLP is available at: \url{https://github.com/fbrausse/smlp}}
%
%
SMLP uses symbolic reasoning for ML model exploration and optimization under verification and stability constraints,
based on SMT, constraint and NN solvers.
In addition its exploration methods are guided by probabilistic and statistical methods.


SMLP is a general purpose tool that requires only data suitable for ML modelling in the csv format (usually samples of the system's input/output). SMLP has been applied at Intel for analyzing and optimizing hardware designs at the analog level.
Currently SMLP supports NNs, polynomial and tree models, and uses SMT solvers for reasoning and optimization at the backend, integration of specialized NN solvers is in progress.
Key algorithms behind SMLP are described in detail in~\cite{DBLP:conf/ijcai/BrausseKK22,DBLP:conf/fmcad/BrausseKK20}.


\SolverAbbrv has been developed by
\href{mailto:franz.brausse@manchester.ac.uk?subject=\SolverAbbrvText}{Franz Brauße},
Zurab Khasidashvili
and Konstantin Korovin and is available
under the terms of the Apache License v2.0.%
\footnote{\url{https://www.apache.org/licenses/LICENSE-2.0}}
\end{abstract}
\tableofcontents

\pagebreak

\section{Introduction}

\emph{Symbolic Machine Learning Prover} (SMLP) offers multiple capabilities for system's \emph{design space exploration}.
These capabilities include methods for selecting which parameters to use in modeling design for configuration optimization and verification;
ensuring that the design is robust against environmental effects and manufacturing variations that are impossible to control, as well as ensuring 
robustness against malicious attacks from an adversary aiming at altering the intended configuration or mode of operation.
Environmental affects like temperature fluctuation, electromagnetic interference, manufacturing variation, and product aging effects are especially 
more critical for correct and optimal operation of devices with analog components, which is currently the main focus area for applying  SMLP. 

To address these challenges, SMLP offers multiple modes of design space exploration; they will be discussed in detail in Section~\ref{sec:exploration}.
The definition of these modes refers to the concept of \emph{stability} of an assignment to system's parameters that satisfies all model constraints 
(which include the constraints defining the model itself and any constraint on model's interface).
We will refer to such a value assignment as a \emph{stable witness}, or \emph{(stable) solution} satisfying the model constraints. 
Informally, stability of a solution means that any eligible assignment in the specified region around the solution also satisfies the required constraints.
This notion is sometimes referred to as robustness. SMLP works with parameterized systems, where parameters (also called \emph{knobs}) can 
be tuned to optimize the system's performance under all legitimate inputs. Parameter optimization under safety constraints is one of the main applications of SMLP.

\begin{figure}[hb!]
\begin{center}
\includegraphics[height=11\baselineskip]{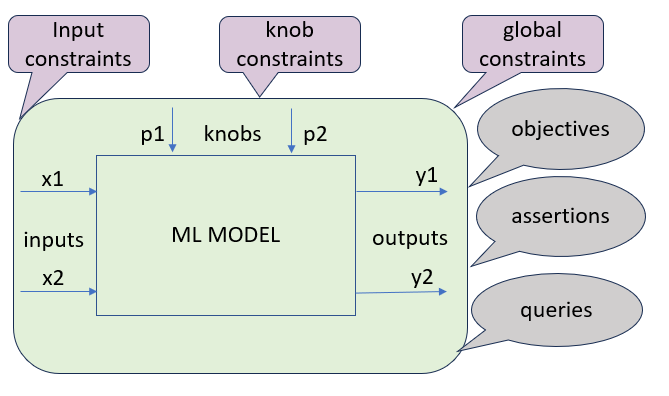}
\caption{Parameterized system, with interface constraints}
\label{fig:system}
\end{center}
\end{figure}

Figure~\ref{fig:system} depicts how SMLP views a system to analyze.
Variables $x1,x2$ are the system's inputs, variables $p1,p2$ are the system's parameters, and variables $y1,y2$ are the system's outputs.
The input, knob, and global constraints on the system's interface define the legal input space of the system as well as requirements that
the system must meet after selecting the knob configuration. The model exploration task might consist of optimizing the system's knobs for 
a number of objectives, synthesizing the knob values to find a witness to a query (e.g., a desired condition), or verifying that a given 
configuration satisfies an assertion on the system's outputs.

For example, in the circuit board design setting, topological layout of circuits, distances, wire thickness, properties of dielectric layers, etc.
can be such parameters, and the exploration goal would be to optimize the system performance under the system's requirements~\cite{9501615}.
The difference between knobs and inputs is that knob values are selected during design phase, before the system goes into operation; on the other 
hand, inputs remain free and get values from the environment during the operation of the system. Knobs and inputs correspond to existentially 
quantified and universally quantified variables in the formal definition of model exploration tasks. Thus in the usual meaning of verification, optimization 
and synthesis, respectively, all variables are inputs, all variables are knobs, and some of the variables are knobs and the rest are inputs.

Below by a \emph{model} we refer to a machine learning model (ML model) that models the system under exploration.

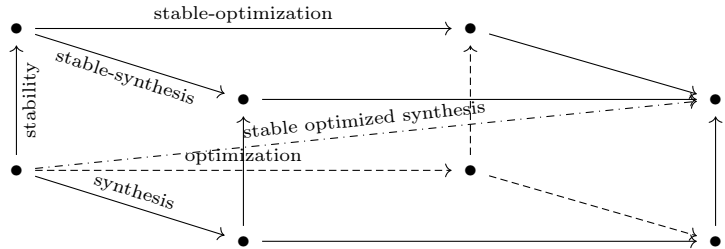
\begin{figure}[hb!]
\centering
\[
    \begin{tikzcd}[row sep=1.5em, column sep = {8.5em, between origins}, my label/.style={midway, sloped}]
    \bullet
    \arrow[rr, "\text{stable-optimization}"]
    \arrow[dr, swap,"\text{stable-synthesis}"{anchor=north,rotate=-18}]
    &&
    \bullet
    \arrow[dr]
    \\
    &
    \bullet
    \arrow[rr]
    &&
    \bullet
    \\
    \bullet
    \arrow[uu, "\text{stability}"{anchor=north,rotate=90}] 
    \arrow[rr, dashed, "\text{optimization}"] 
    \arrow[dr, swap, "\text{synthesis}"{anchor=south, rotate=-18}] 
    \arrow[rrru, dashdotted, "\text{stable optimized synthesis}"{anchor=south, rotate=6}]
    &&
    \bullet  
    \arrow[dr, dashed] 
    \arrow[uu, dashed]
    \\
    &
    \bullet
    \arrow[rr] 
    \arrow[uu]
    && 
    \bullet
    \arrow[uu]
    \end{tikzcd}
\]
\caption{Exploration Cube}
\label{fig:cube}
\end{figure}

The \emph{model exploration cube} in Figure~\ref{fig:cube} provides a high level and intuitive idea on how the model exploration 
modes supported in SMLP are related. The three dimensions in this cube represent synthesis ($\searrow$-axis), optimization 
($\rightarrow$-axis) and stability ($\uparrow$-axis). On the bottom plane of the cube, the edges represent the synthesis and optimization 
problems in the following sense: synthesis with constraints configures the knob values in a way that guarantees that assertions are valid, 
but unlike optimization, does not guarantee optimally with respect to optimization objectives. On the other hand, optimization by itself 
is not aware of assertions on inputs of the system and only guarantees optimality with respect to knobs, and not the validity of assertions 
in the configured system. We refer to the process that combines synthesis with optimization and results in an optimal design that satisfies 
assertions as \emph{optimized synthesis}. The upper plane of the cube represents introducing stability requirements into synthesis 
(and as a special case, into verification), optimization, and optimized synthesis. The formulas that make definition of stable verification,
optimization, synthesis and optimized synthesis precise are discussed Section~\ref{sec:exploration}.

\section{SMLP architecture}

\begin{figure}[tp]
\center
\includegraphics[width= 0.7\columnwidth]{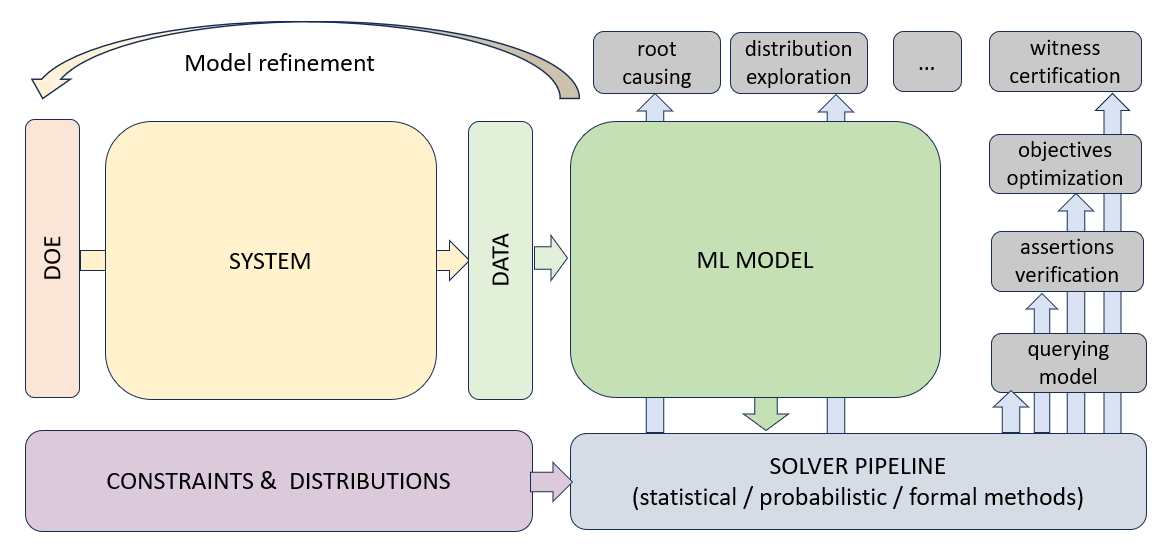}
\caption{SMLP Tool Architecture} \label{smlp_system}
\end{figure}

SMLP tool architecture is depicted in Figure~\ref{smlp_system}. 
It consists of the following components: 1) Design of experiments (DOE),
2) System that can be \changedto{simulated}{sampled} based on DOE, 3)  
ML model trained on the \changedto{simulated}{sampled} data, 4) SMLP solver that handles different system 
exploration modes on a symbolic representation of the ML model, 5) Targeted model refinement loop.

SMLP supports multiple ways to generate training data known under the name of \emph{Design Of Experiments (DOE).}
These methods include: full-factorial, fractional-factorial, Plackett-Burman, Box-Behnken, Box-Wilson, Sukharev-grid, Latin-hypercube, 
among other methods, which try to achieve a smart sampling of the entire input space with a relatively small number of data samples.
In Figure~\ref{smlp_system}, the leftmost box-shaped component called \textsc{doe} represents SMLP capabilities to generate test 
vectors to feed into the system and generate training data; the latter two components are represented with boxes 
called \textsc{system} and \textsc{data}, respectively. 

The component called \textsc{ml model} represents SMLP capabilities to train models; currently neural network, polynomial 
and tree-based  regression models are supported. Modeling analog devices using polynomial models was proposed in the 
seminal work on \emph{Response Surface Methodology (RSM)}~\cite{51647df2-9154-322e-85ba-aa565c9e550a},
and since then has been widely adopted by the industry. Neural networks and tree-based models are used increasingly due to 
their wider adoption, and their exceptional accuracy and simplicity, respectively.

The component called \textsc{solver pipeline} represents model exploration engines of SMLP (e.g., connection to SMT solvers), 
which besides a symbolic representation of the model takes as input several types of constraints and input sampling distributions 
specified on the model's interface; these are represented by the component called \textsc{constraints \& distributions} 
located at the low-left corner of  Figure~\ref{smlp_system}, and will be discussed in more detail in Section~\ref{sec:exploration}.
The remaining components represent the main model exploration capabilities of SMLP. 

Last but not least, the arrow connecting the \textsc{ml model} component back to the \textsc{doe} component represents a
\emph{model refinement} loop which allows to reduce the gap between the model and system responses in the input regions
where it matters for the task at hand (there is no need to achieve a perfect match between the model and the system 
everywhere in the input space). The targeted model refinement loop is discussed in~\Cref{sec:refinement}.

\section{How to run SMLP: a quick start}\label{sec:example}

Command to run SMLP in $\mode{optimize}$ mode is given in \Cref{fig:command}.
Note that all concrete examples in this manual will be executed from the sub-directory \texttt{regr\_smlp/code} of the SMLP distribution.%
\KK{are all files in the repo? if cut \& paste would this run ?}

\begin{figure}[hb!]
\begin{verbatim}
../../src/run_smlp.py -data "../data/smlp_toy_basic" -out_dir ./ -pref Test113 \
-mode optimize -pareto t -resp y1,y2 -feat x1,x2,p1,p2 -model dt_sklearn \
-dt_sklearn_max_depth 15 -mrmr_pred 0 -epsilon 0.05 -delta_rel 0.01 -save_model t \
-model_name test113_model -save_model_config t -plots f -seed 10 -log_time f \
-spec ../specs/smlp_toy_basic.spec
\end{verbatim}
\caption{Example of SMLP's command in mode $\mode{optimize}$, 
to build a decision tree model and perform an optimization task.}
\label{fig:command}
\end{figure}

The option $\optionval{-data}{../smlp\_toy\_basic}$ defines the labeled dataset to use for model training and test.
The dataset should be provided as $\option{smlp\_toy\_basic.csv}$ file;  the $\suffix{.csv}$ suffix itself  may be omitted, zip and bzip2 compressed data files are also accepted.
This dataset is displayed in Table~\ref{toy_basic_df}, and it has six columns $x1, x2, p1, p2, y1, y2$.

The option $\optionval{-mode}{optimize}$ defines the analysis mode to run, and option $\optionval{-pareto}{t}$
instructs SMLP that Pareto optimization should be performed (as opposed to performing multiple single-objective 
optimizations when multiple objectives are specified). 

\begin{table*}[t]
\centering\small
\begin{tabular}{lrrrrrr}
\toprule 
{} &      x1 &  x2 &    p1 &  p2 &       y1 &       y2 \\
\midrule
0 &  2.9800 &  -1 &   0.1 &   4 &   5.0233 &   8.0000 \\
1 &  8.5530 &  -1 &   3.9 &   3 &   0.6936 &  12.0200 \\
2 &  0.5580 &   1 &   2.0 &   4 &   0.6882 &   8.1400 \\
3 &  3.8670 &   0 &   1.1 &   3 &   0.2400 &   8.0000 \\
4 & -0.8218 &   0 &   4.0 &   3 &   0.3240 &   8.0000 \\
5 &  5.2520 &   0 &   4.0 &   5 &   6.0300 &   8.0000 \\
6 &  0.2998 &   1 &   7.1 &   6 &   0.9100 &  10.1250 \\
7 &  7.1750 &   1 &   7.0 &   7 &   0.9600 &   1.1200 \\
8 &  9.5460 &   0 &   7.0 &   6 &  10.7007 &   9.5661 \\
9 & -0.4540 &   1 &  10.0 &   7 &   8.7932 &   6.4015 \\
\bottomrule
\end{tabular}
\caption{Toy dataset $\option{smlp\_toy\_basic.csv}$ with two inputs $x_1, x_2$, two knobs $i_1, i_2$, and two outputs $y_1, y_2$.}
\label{toy_basic_df}
\end{table*}

\begin{figure}
\small
\begin{verbatim}
{
  "version": "1.2",
  "variables": [
    {"label":"y1", "interface":"output", "type":"real"},
    {"label":"y2", "interface":"output", "type":"real"},
    {"label":"x1", "interface":"input", "type":"real", "range":[0,10]},
    {"label":"x2", "interface":"input", "type":"int", "range":[-1,1]},
    {"label":"p1", "interface":"knob", "type":"real", "range":[0,10], "rad-rel":0.1, "grid":[2,4,7]},
    {"label":"p2", "interface":"knob", "type":"int", "range":[3,7], "rad-abs":0.2}
  ],
  "alpha": "p2<5 and x1==10 and x2<12",
  "beta": "y1>=4 and y2>=8",
  "eta": "p1==4 or (p1==8 and p2 > 3)",
  "assertions": {
    "assert1": "(y2**3+p2)/2>6",
    "assert2": "y1>=0",
    "assert3": "y2>0"
  },
  "objectives": {
    "objective1": "(y1+y2)/2",
    "objective2": "y1"
  }
}
\end{verbatim}
\begin{tikzpicture}[remember picture,overlay,shift={(22em,\baselineskip)}]
\node[anchor=south west]{\includegraphics[height=14\baselineskip]{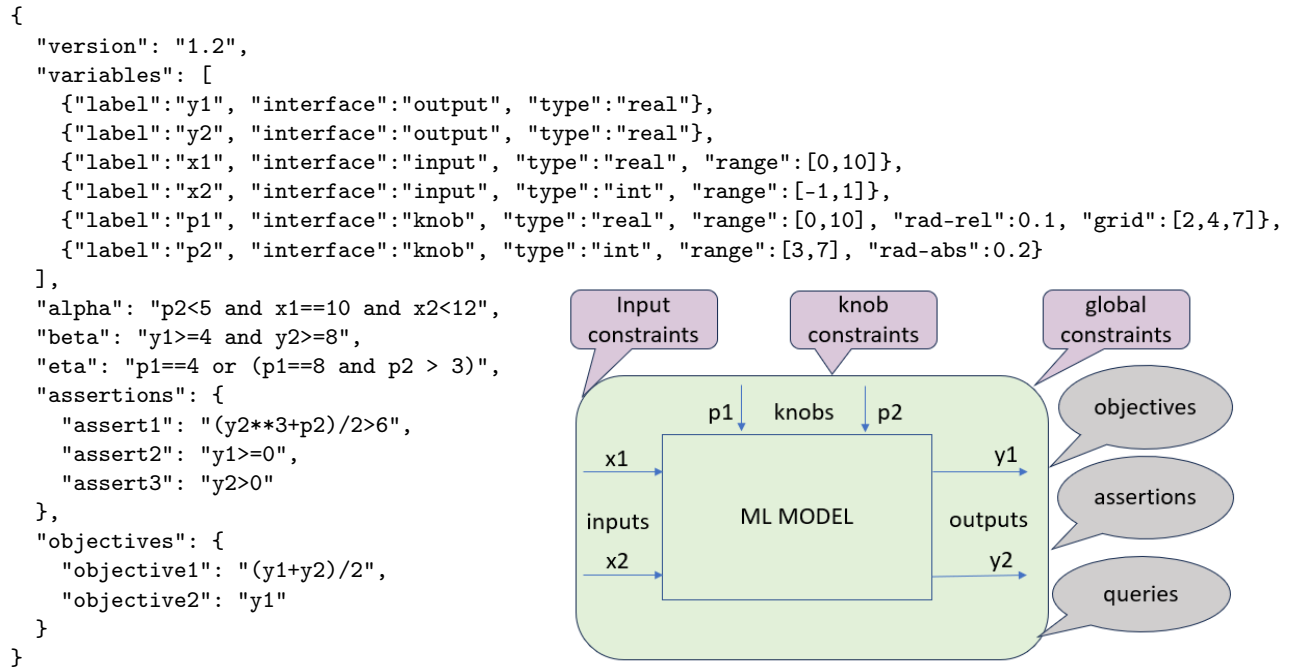}};
\end{tikzpicture}
\vspace*{-1\baselineskip}
\caption{Specification $\option{smlp\_toy\_basic.spec}$ used by SMLP command in Figure~\cref{fig:command}.}
\label{fig:spec}
\end{figure}

The option $\optionval{-spec}{../smlp\_toy\_basic.spec}$ defines the full path to the specification file that 
specifies the optimization problem to be solved.  \Cref{fig:spec} depicts the contents of 
this specification (spec) file. It defines legal ranges of variables $x1, x2, p1, p2, y1, y2$, 
where appropriate, which ones are inputs, which ones are knobs, which ones are the outputs, defines additional 
constraints on them, and defines the optimization objectives. Detailed description of the fields of the specification
(which is loaded as a Python dictionary) is given in Section~\ref{sec:spec}.

Options $\option{-resp\,\,y1,y2\,\,-feat\,\,x1,x2,p1,p2}$ define the names of the responses and features to be used from 
the provided dataset. This information is available in the spec file as well, and therefore these options can be omitted
in our example. In general, option values provided as part of the command line override values of these options
specified in the spec file, and therefore command line options are convenient to quickly adapt an SMLP command
without changing the spec file. Also, a spec file is needed mostly for the model exploration modes of SMLP, and 
command line options make invocation of SMLP in other modes simpler.

Option $\optionval{-model}{dt\_sklearn}$ instructs SMLP to train $\model{dt\_sklearn}$ model, 
which according to SMLP's naming convention for model training algorithms 
means to use the decision tree ($\model{dt}$) algorithm supported in $\package{sklearn}$ package.  
And $\optionval{-dt\_sklearn\_max\_depth}{15}$ instructs SMLP to use the $\package{sklearn}$'s 
$\option{max\_depth}$ hyper-parameter value $\option{15}$, based on a similar naming convention
for hyper-parameters supported in model training packages used in SMLP.

Options $\option{-epsilon \,\, 0.05\,\,-delta\,\,0.01}$ define values for constants $\epsilon$ and $\delta$   
required for approximating search for optima and guaranteeing that the search will terminate. 
These optimization algorithms and proofs that usage of constant $\delta > 0$ guarantees the termination can be found
in~\cite{DBLP:conf/fmcad/BrausseKK20,DBLP:conf/ijcai/BrausseKK22}. Constant $\epsilon$ defines a termination criterion
for search for optima, and is used to guarantee that the computed optima are not more than $\epsilon$ away 
(after scaling the objectives) from the real optima (of the function defined by the ML model).
A formal description of usage of $\epsilon$ can be found in Section~\ref{sec:exploration} as well as 
in~\cite{DBLP:conf/fmcad/BrausseKK20,DBLP:conf/ijcai/BrausseKK22}

Options $\option{-save\_model\,\,t \,\,-save\_model\_config\,\,t}$ instruct SMLP respectively to save the trained 
model and to save the option values used in current SMLP run into a SMLP invocation configuration file. 
Besides saving the model, SMLP saves all required information to enable rerun of the saved model on a new data. 
More details on saving a trained model and reusing it later on a new data is provided in \Cref{sec:models}. 

Option $\optionval{-mrmr\_pred}{0}$ specifies that all features should be used for training a model, while option
value greater than $0$ defines how many features selected by the MRMR algorithm should be used for model training 
(see also \Cref{sec:mrmr:pred}).

After model training (or loading a pre-trained model) SMLP generates plots to visualize model predictions 
against the actual response values found in labeled data ($\option{training|test|new\,\,data}$). 
 Option $\optionval{-plots}{f}$ instructs SMLP to not open these plots interactively while SMLP is running;
these plots are saved for offline inspection. See \Cref{sec:models} for more information regarding prediction plots.

Option $\optionval{-seed}{10}$ is required to ensure determinism in SMLP execution (running the same command 
should yield the same result). And option $\optionval{-log\_time}{f}$ instructs SMLP to not include time stamp in
logged messages.

Option $\optionval{-out\_dir}{../out}$ defines the output directory for all SMLP reports and collateral output files.
See \Cref{sec:smlp:outputs} for more information about SMLP output directory and reports. 

Optimization progress is reported in file $\suffix{try\_smlp\_toy\_basic\_optimization\_progress.csv}$, 
where $\option{try}$ is the run ID specified using option $\optionval{-pref}{try}$; $\suffix{smlp\_toy\_basic}$ 
is the name of the data file, and $\suffix{optimization\_progress.csv}$ is the file name suffix for that report. 
This report contains details on input, knob, output and the objective's values demonstrating the proven 
upper and lower bounds of the objectives during search for a Pareto optimum. It is available anytime
after search for optimum has started and first approximations of the optima have been computed. 
See, Section~\ref{sec:stable:opt} for more details on SMLP reports for mode $\option{optimize}$.
\KK{for later: it would be good to have example with understandable output; sampling from functions?}

\section{Symbolic representation of the ML model exploration}\label{sect.exploration}

The main system exploration tasks handled by SMLP 
can be defined using the GEAR-fragment of $\exists^*\forall^*$ formulas~\cite{DBLP:conf/fmcad/BrausseKK20}:
\begin{equation}\label{form:gear:final}
    \exists p ~\eta(p) \wedge
    \forall p'~
    \forall x y~[
    \theta(p,p') \implies (\varphi_M(p',x,y)  \implies  \varphi_{\mathit{cond}}(p',x,y))
    ]
\end{equation}
where 
$x$ ranges over inputs, $y$ ranges over outputs, and
$p,p'$ range over knobs,
 $\eta(p)$ are constraints on the knob configuration $p$,
  $\varphi_{M}(p',x,y)$ defines the machine learning model, $\theta(p,p')$ defines stability region for the solution $p$, and $\varphi_{\mathit{cond}}(p',x,y)$ defines conditions that should hold in the stability region.


In our formalization $\theta,\eta$ and $\varphi_{\mathit{cond}}$ are quantifier free formulas in the language. These constraints and how they are implemented in SMLP are described below.  


\begin{description}
\item[$\eta(p)$]
    Constraints on values of knobs; 
    this formula need not be a conjunction of constraints on individual knobs, can define more complex relations between allowed knob values of individual knobs. $\eta(p)$ can be specified through the SMLP specification file (see \Cref{sec:spec})
    . 
\item[$\theta(p,p')$]
    Stability constraints that define a region around a candidate solution. 
    This can be specified using either absolute or relative radius $r$ in the 
    specification
    file.
    This region corresponds to a ball (or box) around $p$: $\theta(p,p')\eqdef \Vert p - p' \Vert \leq r$, also denoted as $\theta_r(p)$, in this case was say $p$ is the center point of the region defined by $\theta_r(p)$. In general, our methods do not impose any restrictions on $\theta$ apart from reflexivity.
\item[$\varphi_M(p,x,y)$]
    Constraints that define the function represented by the ML model $M$, thus
    $\varphi_M(p,x,y) \eqdef(M(p,x) = y)$.  
    In the ML model knobs are represented as designated inputs (and can be treated in the same way as system inputs, or the machine model architecture can reflect the difference between inputs and knobs).
    $\varphi_M(p,x,y)$ is computed by SMLP internally, based on the ML model specification.
\item[$\varphi_{\mathit{cond}}(p,x,y)$]    
    Conditions that should hold in the $\theta$-region of the  solution.
    These conditions depend on the exploration mode and could be:
    (1) verification conditions,
    (2) model querying conditions, 
    (3) parameter optimization conditions, or
    (4) parameter synthesis conditions.
    The exploration modes are described in 
    \Cref{sec:exploration}.
\end{description}

SMLP solver 
is based on specialized procedures for solving formulas in the GEAR fragment using quantifier-free SMT solvers, GearSAT$_\delta$~\cite{DBLP:conf/fmcad/BrausseKK20} and GearSAT$_\delta$-BO~\cite{DBLP:conf/ijcai/BrausseKK22}.
The GearSAT$_\delta$ procedure interleaves search for candidate solutions using SMT solvers with exclusion of $\theta$-regions around counterexamples. GearSAT$_\delta$-BO combines GearSAT$_\delta$ search with Bayesian optimization guidance.
These procedures find
solutions to GEAR formulas
with user-defined accuracy $\varepsilon$ 
and they have been proven to be sound, ($\delta$)-complete and terminating%
.

\section{SMLP problem specification}\label{sec:spec}

The specification file defines the problem conditions in a JSON compatible format, whereas SMLP exploration modes 
can be specified via command line options. \Cref{fig:spec} depicts a toy system with two inputs, two knobs, 
and two outputs and a matching specification file for model exploration modes in SMLP. This system and the spec file 
were used in \Cref{sec:example} to give a quick introduction on how to run SMLP in the $\mode{optimize}$ mode. \KK{which fields are optional ? might be use (?) or (-)/(+) for optional/required fields, or just mention in text}

\begin{itemize}
\item[$\speckey{version}$] specifies the version of the spec file format. Versions are defined for backward compatibility.
\item[$\speckey{variables}$] defines properties of the system's interface variables. For each variable it specifies its
\begin{itemize}
\item[$\speckey{label}$] the name, e.g., $x1$.
\item[$\speckey{interface}$]  function, which can be $\dtype{input}$, $\dtype{knob}$,  or $\dtype{output}$ .
\item[$\speckey{type}$] which can be $\dtype{real}$, $\dtype{int}$, or $\dtype{set}$ (for categorical features).
\item[$\speckey{range}$]  for variables of $\dtype{real}$ and $\dtype{int}$ types, e.g., $\specval{[2,4]}$  
(must be a closed interval). Values $\specval{-\inf}$ and $\specval{\inf}$ are allowed as the min and max of the 
range, and can be specified using $\specval{null}$. e.g. $\specval{[-2, null]}$.
\ZK{Implementation allows plus/minus infinity -- discourage its usage because of performance considerations?}
The input and knob ranges serve as assumptions in model exploration modes of SMLP. 
\ZK{Need to be clear about ranges of outputs -- what is their meaning and usage, if specified?}
\item[$\speckey{rad-abs}$] absolute stability radius for knobs. 
\item[$\speckey{rad-rel}$] relative (wrt, to the center point of the region) stability radius for knobs. Only one type of radii is needed per parameter.
\item[$\speckey{grid}$] for knobs, which is a list of values that a knob variable is allowed to take within the respective declared ranges, 
independently from other knobs. The $\eta$ constraints introduced below further restrict the multi-dimensional grid. 
Both $\dtype{real}$ and $\dtype{int}$ typed knobs can be restricted to grids (but do not need to).
Grids serve as assumptions in model exploration modes of SMLP.
\end{itemize}
\item[$\speckey{eta}$] defines extra constraints on knobs
(on top of constraints inferred from knob ranges and grids).
\item[$\speckey{alpha}$] defines extra constraints on inputs and knobs 
(on top of constraints inferred from input and knob ranges and knob grids).
These constraints serve as assumptions in model exploration modes of SMLP. 
\item[$\speckey{beta}$] defines constraints on inputs, knobs and outputs
that serve as requirements that need to be met by selected knob configurations. 
\item[$\speckey{assertions}$] defines assertions: a dictionary that maps assertion names to respective expressions. 
\item[$\speckey{queries}$] defines queries: a dictionary that maps query names to respective expressions.
\item[$\speckey{objectives}$] defines optimization objectives: a dictionary that maps objective names to respective expressions.
\end{itemize}

The expressions that occur in a spec file, such as $\speckey{alpha}$, $\speckey{beta}$, $\speckey{eta}$ constraints,
as well as $\speckey{assertions}$, $\speckey{queries}$, and $\speckey{objectives}$, can in principle be any Python expression
that can be composed using the $\package{operator}$ package\footnote{\url{https://docs.python.org/3/library/operator.html}}. These constraints are formally introduced in Section~\ref{sec:exploration}.

In SMLP these expressions are parsed using the 
$\package{Abstract\,\,Syntax\,\,Trees}$ library\footnote{\url{https://docs.python.org/3/library/ast.html}}.
Currently only a subset of operations from the $\package{operator}$ package is supported in these expressions 
(there has not been a need for others so far):
\begin{itemize}
\item[binaryop] $\operator{add(a,b)\,\,[a+b]}$, $\operator{sub(a,b)\,\,[a-b]}$, $\operator{mul(a,b)\,\,[a*b]}$, 
$\operator{truediv(a,b)\,\,[a/b]}$, $\operator{pow(a,b)\,\,[a**b]}$
\item[unaryop] $\operator{neg(a)\,\,[-a]}$
\item[bitwiseop] $\operator{and\_(a,b)\,\,[a\&b]}$, $\operator{or\_(a,b)\,\,[a|b]}$, $\operator{inv(a)\,\,[\sim a]}$, 
$\operator{xor(a,b)\,\,[a\,\,\hat{}\,\,b]}$
\item[cmpop] $\operator{eq(a,b)\,\,[a == b]}$, 
$\operator{ne(a,b)\,\,[a != b]}$, $\operator{lt(a,b)\,\,[a<b]}$, $\operator{le(a,b]\,\,[a\leq b]}$, $\operator{gt(a,b)\,\,[a>b]}$, 
$\operator{ge(a,b)\,\,[a\geq b]}$ 
\item[if-then-else] $\operator{x\,\, if\,\,cond\,\,else\,\, y\,\,[Ite(cond, x, y)]}$
\end{itemize}

The $\speckey{alpha}$, $\speckey{beta}$, $\speckey{eta}$ constraints can also be defined in SMLP 
command line using options $\option{-alpha\,\,expression, -beta\,\,expression, -eta\,\,expression}$.
Assertions can be specified as part of command line, using options $\option{-asrt\_names}$ and $\option{-asrt\_exprs}$.
For example, assertions from the spec in Figure~\ref{fig:spec}
can be specified as follows: $\option{-asrt\_names\,\,assert1,assert2,assert3}$ 
specifies assertion names as a comma-separated list of names, and 
$\option{-asrt\_expr\,\,"(y2**3+p2)/2>6;y1>=0;y2>0"}$
defines the respective expressions as a semicolon separated list of expressions.
Similarly, queries can be specified in command line using options 
$\option{-quer\_names}$ and $\option{-quer\_exprs}$; and optimization objectives can be
specified using options $\option{-objv\_names}$ and $\option{-objv\_exprs}$.

Precise handling of constraints $\speckey{alpha}$, $\speckey{beta}$, $\speckey{eta}$, as well as handling of
 $\speckey{assertions}$,  $\speckey{queries}$,  and $\speckey{objectives}$, depends on the model exploration 
modes of SMLP and is described in dedicated subsections of Section~\ref{sec:exploration}.

\section{SMLP input and output}\label{sec:smlp:inputs:outputs}

Input files to an SMLP command can be located in different directories and have in general different formats.
The most common input files and how to feed them to SMLP is described in Subsection~\ref{sec:smlp:inputs}. 
All outputs from an SMLP run, on the other hand, are written into the same output directory, as described in 
Subsection~\ref{sec:smlp:outputs}. 

\subsection{SMLP inputs}\label{sec:smlp:inputs}

\begin{itemize}
\item[training data] should be a $\suffix{.csv}$ file, possibly compressed as $\suffix{.csv.bz2}$ or $\suffix{.csv.gz}$. 
Full or relative path to data should be specified using option $\option{-data}$; the $\suffix{.csv}$ suffix can be omitted
in case the data file is not compressed. For modes where a model is trained, data should include one or more responses.
All responses must be numeric (categorical features as responses will be supported in future). 
The data file is relevant for all modes of SMLP except for the $\mode{doe}$ mode.
\item [new data] should be a $\suffix{.csv}$ file, possibly compressed as $\suffix{.csv.bz2}$ or $\suffix{.csv.gz}$. 
Full or relative path to data should be specified using option $\option{-new\_data}$; the $\suffix{.csv}$ suffix can be omitted
in case the data file is not compressed. This data usually is not available during model training, and usually is also not labeled:
it may not contain the response columns and should contain all features from the training data that were actually used in model training.
New data is used to perform prediction with a model trained on training data; this model can be generated in the same SMLP
run or could have been trained and saved earlier. New data is mainly relevant for mode $\mode{predict}$, but new data
ca be supplied in model exploration modes as well and in this case predictions on new data will be performed as part of
model exploration analysis. When new data has the responses, they must be of the same type as in the training data, 
and after predictions the model accuracy will be reported for both training and new data.
\item [problem spec] should be a $\suffix{.spec}$ file, with the content in $\suffix{json}$ format (so it is loaded using json.load() 
as a Python dictionary). Full or relative path to spec file, including the $\suffix{.spec}$ suffix, should be specified using option 
$\option{-spec}$. It is required in model exploration modes 
($\mode{certify}$, $\mode{query}$, $\mode{verify}$, $\mode{synthesize}$, $\mode{optimize}$, $\mode{optsyn}$).
\item[doe spec] should be a $\suffix{.csv}$ file. Full or relative path to DOE (design of Experiments) spec file should be specified 
using option $\option{-doe\_spec}$. It is required for the $\mode{doe}$ mode only, for DOE generation.
\end{itemize}

\subsection{SMLP outputs}\label{sec:smlp:outputs}

SMLP communicates its results using files, and it outputs all reports, plots, and collateral files in the same directory.
A full path to that output directory can be specified using option $\option{-out\_dir}$, and it is recommended to specify it.
If not specified, the directory of input data file is used as the output directory. 
If the latter is not specified (say if a saved model is used for performing prediction), the directory of the new data 
file is used as the output directory. If the new data file is not specified either, say in case of $\mode{doe}$ mode, 
then the directory of the DOE spec file is used as the output directory. Otherwise an error is issued.

The output files may also include a saved trained model and a collection of other files that together have all the information
required to rerun the saved model on new data. All files collectively defining a saved model start with the same name prefix.
This prefix is a concatenation of the SMLP invocation ID/name specified using option $\option{-pref\,\,runname}$, and the
saved model name specified using option $\option{-model\_name\,\, modelname}$. If saved model name is not specified,
the prefix for all model related file names is computed by SMLP using the data name that was used for training the model,
but this might change in future and it is recommended to always use a model name when saving a trained model. 

All the other output file names also have the same prefix, computed by concatenating the SMLP run ID/name 
specified using option  $\option{-pref}$ and the name of input data (or new data) file in modes where these
data files are provided, or with the name of the saved model if the latter is used in analysis, or the DOE spec file
name in the $\mode{doe}$ mode. Currently any SMLP mode uses at least one of the following: input (training)
data, new data, or DOE spec file, therefore file name prefixes are well defined both for saved model related files
as well as SMLP report and collateral files. Assuming a unique ID/name is used for each SMLP run (specified
using option $\option{-pref}$), all files generated as a result of that run can be identified uniquely.

\section{Data processing options}

In SMLP we distinguish between two stages of input data processing: a \emph{data preprocessing stage}, 
followed by a \emph{data preparation stage} for the required type of analysis.

\subsection{Data preprocessing options}

Data \emph{preprocessing} is applied to raw data immediately after loading, and its aim is to process
data in order to confirm to SMLP data requirements. That is, this
stage of data processing is to make SMLP tool user friendly, and perform some data 
transformations instead of the user having to do this. Thus, all the reports and visualization
of the results will use preprocessed data, and assume the data was passed to SMLP in that
form. As an example, if some values in columns were replaced in the preprocessing stage, 
say 'pass' was replaced by $0$ and 'fail' was replaced by $1$, the reports will use values 
$0$ and $1$ in that column. 

Next we explain the main steps performed as part of preprocessing of training data.

\subsubsection{Selecting features for analysis} 

If SMLP command includes option $\option{-feat\,\, x,y,z}$, then only features $x,y,z$ will be used
in analysis (besides the responses); the rest of the features will be dropped.

\subsubsection{Missing values in responses}  

Response columns, say $y1, y2$, in training data are defined using option $\option{-resp \,\,y1. y2}$.
Rows in the training data where at least one response has a missing value will be dropped.

\subsubsection{Constant features} 

Constant features (that have exactly one non-NaN value) are dropped.

\subsubsection{Missing values in features} 

Missing value imputation is performed with the $\option{most\_frequent}$ strategy of 
$\module{SimpleImputer}$ class from $\package{sklearn}$ package. The locations of missing values
prior to imputation is computed as a dictionary and saved as a json file with suffix $\suffix{\_missing\_values\_dict.json}$
for future reference (say to mark respective locations or samples on plots).

\subsubsection{Boolean typed features} 

Currently SMLP does not have a need to make a direct usage of boolean type in features (or in responses). 
Therefore Boolean typed features are treated as categorical  features with type $\dtype{object}$, by converting 
the Boolean values to $\dtype{strings}$ 'True' and 'False'.

\subsubsection{Determining types of responses}

\noindent \emph{Categorical responses:} 
Categorical responses are supported only if they have two values -- it is user 
responsibility to encode a categorical response with more than two levels (values) into
a number of binary responses (say through the one-hot encoding). 
A categorical response can be specified as a (a) 0/1 feature, (b) categorical feature with 
two levels; or (c) numeric feature with two values. In all cases, parameters specified through
options $\option{positive\_value}$ and $\option{negative\_value}$ determine which one of 
these two values in that response define the positive samples and which ones define the 
negative ones -- both in training data and in new data if the latter has that response column. 
Then, as part of data preprocessing, the  $\option{positive\_value}$ and the $\option{negative\_value}$ 
in the response will be replaced by $1$ and $0$, respectively, following the convention in statistics that 
integer $1$ denotes positive and $0$ denotes negative. 
\delete{
That is, the intention is that user has to provide categorical responses as $1/0$ responses
where $1$ denotes positive samples and $0$ denotes negative samples, and to make the tool user
friendly the user has freedom to specify categorical responses in one of the three ways (a)-(c)
described above with the intention that the tool internally will convert such responses into
$1/0$ columns and all the results will be reported and visualized using $1/0$ values and not the
original values used to define positive and negative values in the responses in the raw input data.
}

\noindent \emph{Numeric response columns:} 
Float and int columns in input data can define numeric responses. Each such response with more than 
two values is treated as numeric (and we are dealing with a regression analysis). If a response has 
two values, than it can still be treated as a categorical/binary response, as described in case (c) of specifying
binary responses. Otherwise -- that is, when $\{\option{positive\_value}, \option{negative\_value}\}$ is not
equal to the set of the two values in the response, the response is treated as numeric. 
\delete{
Parameter values specified through options $positive\_value$ and $negative\_value$ have a different 
meaning for numeric responses: they are not used to replace values in the response as part of 
preprocessing. Instead, $positive\_value = STAT\_POSITIVE\_VALUE$ and $negative\_value = STAT\_NEGATIVE\_VALUE$
(which is the default) specifies that the high values in the response are positive (undesirable) and
the low values are negative (desirable). The opposite assignment $positive\_value$ = $STAT\_NEGATIVE\_VALUE$ 
and $negative\_value = STAT\_POSITIVE\_VALUE$ specifies that low values in the response are positive and
high values are negative. Other possibilities for the pair $(positive\_value, negative\_value)$ are 
considered as incorrect specification and an error message is issued. 
}

\noindent \emph{Multiple responses}: 
Multiple responses can be treated in a single SMLP run only if all of them are identified as 
defining regression analysis or all of them are identified as defining classification analysis.
if that is not the case, SMLP will abort with an error message clarifying the reason.

\delete{
\subsection{Optimization problems}

If we are dealing with an optimization problem for a response or multiple responses, then combination
$positive\_value = STAT\_POSITIVE\_VALUE$ and $negative\_value = STAT\_NEGATIVE\_VALUE$ specifies that 
we want to maximize the response values (find regions in input space where the responses are
close to maximum / close to Pareto optimal with respect to maximization; and conversely, combination
$positive\_value = STAT\_NEGATIVE\_VALUE$ and $negative\_value = STAT\_POSITIVE\_VALUE$ 
specifies that we are looking at (Pareto) optimization problem with respect to minimization.
}

\subsection{Data preparation for analysis}

We now describe data preparation steps supported in SMLP.

\subsubsection{Processing categorical features}

After preprocessing the only supported (and expected) data column types are $\dtype{int}$,  
$\dtype{float}$ and $\dtype{categorical}$, where categorical features can have types  $\dtype{object}$
(with values of type $\dtype{string}$), or $\dtype{category}$; the  $\dtype{category}$ type can be 
 $\dtype{ordered}$ or  $\dtype{unordered}$.

Some of the ML algorithms prefer to use categorical features as is -- with string values: for example,
feature selection algorithms can use dedicated correlation measures for categorical features. Also,
some of the model training algorithms, such as tree based, can deal with categorical features directly, 
while others, e.g., neural networks and polynomial models, assume all inputs are numeric
($\dtype{int}$ or $\dtype{float}$). Therefore, depending on the analysis mode (feature selection, model training,
model exploration), categorical features might be encoded into integers (and be treated as discrete 
domains), simply by enumerating the levels (the values) seen in categorical
features and replacing occurrences of each level with the corresponding integer. Currently 
encoding categorical features as integers is the default in model training and exploration modes in SMLP.

Conversely, some ML algorithms (especially, correlations)
might prefer to \emph{discretize} numeric features into categorical features, and discretization options
in SMLP support discretization of numeric features with target types $\dtype{object}$ and $\dtype{category}$, 
 $\dtype{ordered}$ or  $\dtype{unordered}$, where the values in the resulting columns can represent integers 
(as strings,  e.g.,  '5', or as levels, e.g., $5$), or other string values (like 'bin5'). Discretization is controlled
using the following options: 
\begin{itemize}
\item $\option{discr\_algo}$: discretization algorithm can be $\option{uniform}$, $\option{quantile}$, 
$\option{kmeans}$, $\option{jenks}$, $\option{ordinals}$, $\option{ranks}$.
\item $\option{discr\_bins}$: specifies number of required bins.
\item $\option{discr\_labels}$: if true, string labels (e.g., 'Bin2') will be used to denote levels of the categorical feature
resulting from discretization; otherwise integers (e.g., 2) will be used to represent the levels.
\item $\option{discr\_type}$: the resulting type of the obtained categorical feature; can be specified as 
$\dtype{object}, \dtype{category}, \dtype{ordered}$, and $\dtype{integer}$.
\end{itemize}

\subsubsection{Feature selection for model training}\label{sec:mrmr:pred}

SMLP incorporates the MRMR feature selection algorithm~\cite{DBLP:journals/jbcb/DingP05} 
for selecting a subset of features that will be used for model training, using Python package 
$\package{mrmr}.$\footnote{\url{https://github.com/nlhepler/mrmr}.}
SMLP option $\option{-mrmr\_pred\,\,15}$ instructs the MRMR algorithm to select $15$ features, 
according to the principle of \emph{maximum relevance and minimum redundancy}.

\subsubsection{Data scaling / normalization}

Data scaling is managed separately for the features and the responses. A particular mode of usage
(model training and prediction, feature selection or subgroup discovery, Pareto optimization, etc.)
can decide to scale features and or scale responses. The reports and visualization should use
features and responses in the original scale, thus unscaling must be performed. 

Features and responses might or might not be scaled, and these are controlled using two options:
the option $\option{-data\_scaler}$  controls which data scaler should be used: the $\option{MinMaxScaler}$ 
class of $\package{sklearn}$ package or $\option{none}$  (in which case neither features nor responses can be scaled); 
and Boolean typed options $\option{-scale\_feat}$ and $\option{-scale\_resp}$ for controlling feature and 
response scaling, respectively.

SMLP optimization algorithms operate with data in original scale, while the optimization objectives are scaled 
(always,  in current implementation) to $[0,1]$ based on the min and max values each individual objective function takes on samples
in the training data.

\subsection{Processing of new data}

In model training and exploration modes, most of the above described data processing steps are applied to 
training data. New data for performing predictions, if supplied, requires related feature processing and sanity 
checks to ensure that it does not contain any features not used in model training, and categorical features
in new data do not have levels that were not present in the same features in training data.
Some processing steps, such as missing value imputation in features, are applied to both training and
new data.

\subsection{Output files during data processing}\label{sec:data:outputs}

The following information is computed and saved in output files during data processing stages.
This information is required for performing predictions based on a saved model as well as in
model exploration modes.

\begin{itemize}
\item $\suffix{*\_data\_bounds.json}$: Dictionary, with feature and response names in training data as the dictionary keys; 
and the min/max info of these features and responses as the dictionary values. 
\item $\suffix{*\_missing\_values\_dict.json}$: Dictionary, with names of features that have at least one missing value as the 
dictionary keys; and the list of indices of missing values in these features  as the dictionary values.
\item $\suffix{*\_model\_levels\_dict.json}$: Dictionary, with names of categorical features in input data as the dictionary 
keys, and the levels (the values) in these features as the dictionary values.
\item $\suffix{*\_model\_features\_dict.json}$: Dictionary, with names of responses as the dictionary keys; and the names of
features used to train model for that response as the dictionary values.
\item $\suffix{*\_features\_scaler.pkl}$: Object of $\option{MinMaxScaler}$ clas from $\package{sklearn}$ package, 
used for scaling features, saved as $\suffix{.pkl}$ file.
\item $\suffix{*\_responses\_scaler.pkl}$:  Object of $\option{MinMaxScaler}$ clas from $\package{sklearn}$ package, 
used for features scaling the responses, saved as $\suffix{.pkl}$ file. 
\end{itemize}

\section{ML model training and prediction}\label{sec:models}

In this section we describe SMLP modes $\mode{train}$ and $\mode{predict}$: 
how to train ML models with SMLP, how to save them, and how to rerun saved models on new data are decried 
in \Cref{sec:models:train}, \Cref{sec:models:save}, and \Cref{sec:models:rerun}, respectively. Reports and other 
collateral files generated in $\mode{train}$ and $\mode{predict}$ modes is discussed in~\Cref{sec:models:reports}.

\subsection{Training ML models}\label{sec:models:train}

SMLP supports training tree-based and polynomial models using the $\package{scikit-learn}$\footnote{\url{https://scikit-learn.org/stable/}} 
and $\package{pycaret}$\footnote{\url{https://pycaret.org}} packages, and training neural networks using the $\package{Keras}$ package 
with $\package{TensorFlow}$\footnote{\url{https://keras.io}}. For systems with multiple outputs (responses), SMLP supports training 
one model with multiple responses as well as training separate models per response  (this is controlled by command-line option 
$\option{-model\_per\_response}$). Supporting these two options allows a trade-off between the accuracy 
of the models (models trained per response are likely to be more accurate) and with the size of the formulas that represent the model for 
symbolic analysis (one multi-response model formula will be smaller at least when the same training hyper-parameters are used). 
Conversion of models to formulas into SMLP language is done internally in SMLP (no encoding options are exposed to user in current 
implementation, which will change once alternative encodings will be developed).

Figure~\ref{fig:pred:command} displays an example SMLP command in $\mode{prediction}$ mode.
The option $\option{-data}$ specifies full path to data $\suffix{csv}$ file (the $\suffix{.csv}$ suffix is not required).
Similarly, option $\option{-new\_data}$ specifies full path to the new data (usually not available/used during model training and validation).
Option $\option{-resp}$ defines the names of the responses;
and option $\option{-feat}$ defines the names of a subset of features from training data to be used in ML model training
(the same subset of features is selected form new data to perform prediction on new data).
Option $\option{-model}$ defines the ML model training algorithm. As an example, the command in Figure~\ref{fig:pred:command}
trains a polynomial model using the $\package{scikit-learn}$ package, and in SMLP model training algorithm naming convention is to
suffix the algorithm name with the package name, separated by underscore,  to form the full algorithm name.
For $\package{Keras}$ and $\package{pycaret}$ packages, the package name suffixes used are $\package{keras}$ and 
$\package{pycaret}$, respectively, while for algorithms from $\package{scikit-learn}$ package we use an 
abbreviated suffix $\package{sklearn}$.
\begin{figure}
\begin{verbatim}
../src/run_smlp.py -data ../smlp_toy_basic -out_dir ../out -pref test_predict \
-mode predict -resp y1,y2 -feat x1,x2,p1,p2 -model poly_sklearn -save_model t \ 
-model_name test_predict_model -save_model_config t -mrmr_pred 0 -plots f \
-seed 10 -log_time f -new_data ../smlp_toy_basic_pred_unlabeled  

\end{verbatim}
\caption{Example of SMLP's command  to train a polynomial model and perform prediction on new data.}
\label{fig:pred:command}
\end{figure}

SMLP command for mode $\mode{train}$ is similar: the mode is specified using $\optionval{-mode}{train}$;
and specifying new data with option $\option{-new\_data}$ is not required.

\subsection{Saving ML models}\label{sec:models:save}

SMLP supports saving a trained model to reuse it in future on incoming new data. 
Saving a model is enabled using options $\option{-save\_model\,\, t\,\,-model\_name \,\,modelname}$,
and rerunning a saved model is enabled using options $\option{-use\_model\,\, t\ \,\,-model\_name \,\,../modelname}$.
In addition, using options $\option{-save\_model\,\, t\,\,-model\_name \,\,modelname\,\, -save\_config\,\, t}$ 
one can generate a configuration file that records model training options prior to saving the model and  enables one 
to easily rerun the saved model. The $\option{nn\_keras}$ models are saved and loaded using the $\suffix{.h5}$
format, while for models trained using $\package{sklearn}$ and $\package{pycaret}$ packages the  $\option{.pkl}$ format 
 is used.

\subsection{Rerunning ML models}\label{sec:models:rerun}

Figure~\ref{fig:pred:rerun:saved} gives two example commands to rerun a saved model on new data.
The first one repeats the options of the command that trained and saved the model, and the second
one uses the configuration file saved during the model training (the configuration file records all the SMLP options
used during model training; therefore there is no need to repeat the SMLP options used during model training
when reusing the saved model with the configuration file). 

\begin{figure}
\begin{verbatim}
../src/run_smlp.py -model_name ../test_predict_model -out_dir ../out \ 
-pref test_prediction_rerun -new_data ../smlp_toy_basic_pred_unlabeled \
-config ../test_predict_model_rerun_model_config.json 

../src/run_smlp.py -mode predict -resp y1,y2 -feat x1,x2,p1,p2 -out_dir ../out \ 
-use_model t -model_name ../test_predict_model -model poly_sklearn \
-save_model f -pref model_rerun -mrmr_pred 0 -plots f \
-seed 10 -log_time f -new_data ../smlp_toy_basic_pred_unlabeled

\end{verbatim}
\caption{Examples of SMLP's commands to use a saved mode to perform prediction on new data.}
\label{fig:pred:rerun:saved}
\end{figure}

\subsection{ML model training and prediction reports}\label{sec:models:reports}

Here is the list of reports and collateral files generated in SMLP modes that require ML model training or rerunning 
of a saved ML model. Recall that new data is available in mode $\mode{predict}$ and is not available in mode 
$\mode{train}$, and may or may not be available in model exploration modes.

\begin{itemize}
\item $\suffix{*\_\{training|test|labeled|new\}\_predictions\_summary.csv}$: prediction results respectively on 
training data samples, on test data (also called validation data) samples, on the entire labeled data samples
(which includes both training and test data samples), and on new data samples (when available). It is saved as 
a $\suffix{.csv}$ file that includes the values of the responses as well.
\item $\suffix{*\_\{training|test|labeled|new\}\_prediction\_precisions.csv}$: prediction previsions per response, 
respectively on training data samples, on test data (also called validation data) samples, on the entire labeled data 
samples (which includes both training and test data samples), and on new data samples (when available). It is saved 
as a $\suffix{.csv}$ file. For regression models currently supported 
in SMLP, two measures of precision are reported: $\precision{msqe}$, and $\precision{r2\_score}$.
\item $\suffix{*\_\{training|test|labeled|new\}\_\{dt\_sklearn|poly\_sklearn|nn\_keras|\ldots\}.png}$: 
response value distribution and prediction plots that display real vs predicted values respectively
for training, test, labeled and new data (when available), for the model trained respectively using 
$\model{dt\_sklearn}$, $\model{poly\_sklearn}$, $\model{nn\_keras}$, 
or other regression model training algorithms supported in SMLP. 
Generation of response value distribution plots and prediction accuracy plots
are controlled using options $\option{-resp\_plots\,\,t}$ and $\option{-pred\_plots\,\,t}$, respectively.
When generated, these plots are saved for an offline review, and can also be displayed during SMLP
execution, for an interactive review, using option $\option{-plots\,\,t}$. 
\item $\suffix{*\_\{dt\_sklearn|poly\_sklearn|nn\_keras|\ldots\}\_model\_complete.\{h5|pk\}}$: saved mode
in $\suffix{.h5}$ format for $\model{nn\_keras}$ and in  $\suffix{.pkl}$ format for ML models trained 
using packages $\package{sklearn}$ and $\package{caret}$.
\item $\suffix{*\_rerun\_model\_config.json}$: SMLP options configuration file created when saving 
a trained model, and loaded when re-using the saved model. This configuration file records all option values in the 
SMLP run that trains the model, and it can be used to rerun the model on a new data using option 
$\option{-config\,\,full\_path\_to\_config\_file.json}$, as described earlier in this section. During rerun, usually
options $\option{-pref}$, $\option{-new\_data}$ and $\option{-model\_name}$, with full paths to new data 
and saved model files (with the model name as prefix), respectively, are specified along with the configuration file, 
and these option values override the respective option values recorded within the configuration file.
Besides the saved model itself, rerunning a saved model requires several other files saved during data processing
steps (preprocessing and data preparation for analysis), which are described in Section~\ref{sec:data:outputs}.  
\end{itemize}

\section{ML model exploration with SMLP}\label{sec:exploration}

SMLP supports the following model exploration modes (we assume that an ML model $M$ has already been trained). 
Precise descriptions of these modes are given in the subsequent subsections. \KK{cmd line options}

\begin{itemize}
\item[certify] Given an ML model $M$, a value assignment $p^*,x^*$ to knobs $p$ and to inputs $x$, and a query $\query(p,x,y)$,
check that $p^*,x^*$ is a stable witness to $\query(p,x,y)$ on model $M$. Multiple pairs $(\specval{witness},\query)$ 
of candidate witness $\specval{witness}$ and query $\query$ can be checked in a single SMLP run.
\item[query] Given an ML model $M$  and a query $\query(p,x,y)$, find a value assignment $p^*,x^*$ to knobs $p$ and to inputs $x$
that serves as a stable witness for $\query(p,x,y)$ on $M$. Multiple queries can be evaluated in a single SMLP run.
\item[verify] Given an ML model $M$,  a configuration $p^*$ of knobs $p$ (that is, a value assignment $p^*$ to knobs $p$),  
and an assertion $\assert(p,x,y)$, verify whether  
$\assert(p',x,y)$ is valid on model $M$ for any assignment $p'$ to knobs in the stability region of $p^*$ and all
legal values of inputs $x$.  SMLP supports verifying multiple assertions in a single run.
\item[synthesize] Given an ML model $M$,  find a configuration $p^*$ of knobs $p$
such that all required constraints, including assertions,  are valid for any configuration $p'$ of knobs in the 
stability region of $p^*$  and any legal values of inputs $x$.
\item[optimize]  Given an ML model $M$,  find a stable configuration $p^*$ of knobs $p$ that yields a Pareto-optimal values of 
    the optimization objectives
    (Pareto-optimal with respect to the \emph{max-min} optimization problem defined 
in~\cite{brausse2024smlp}).
\item[optsyn] Given an ML model $M$,  find a configuration $p^*$ of knobs $p$ that yields a Pareto-optimal values of the 
optimization objectives and such that all constraints and assertions are valid for any configuration $p'$ of knobs in the 
stability region of $p^*$  and legal values of inputs $x$. This mode is a union 
of the \emph{optimize} and \emph{synthesize} modes, its full name is \emph{optimized synthesis}. 

\end{itemize}

Table~\ref{smlp:modes:features} summarizes components relevant to verification, synthesis and optimization
in their regular meaning,  and the above listed model exploration modes in SMLP. These features include
relevance of inputs, knobs, stability, constraints, as well as  queries, assertions and objectives,
to particular model exploration modes in SMLP. 
Algorithms for all other modes can be seen as a sub-procedures of optimized synthesis algorithm.

\begin{table}
 \begin{tabular}{lllllllllll}
\toprule
           mode / feature & inputs & knobs & $\alpha$ & $\beta$ & $\eta$ & $\theta$ & $\delta/\epsilon$ & queries & assertions & objectives \\
\midrule
   verification &    yes &    no &   yes &   no &  no &    no &            no &      no &        yes &         no \\
      synthesis &    yes &   yes &   yes &  yes & yes &    no &            no &      no &        yes &         no \\
   optimization &     no &   yes &   yes &  yes & yes &    no &            no &      no &         no &        yes \\
   SMLP-certify &    yes &   yes &   yes &   no & yes &   yes &            no &     yes &         no &         no \\
     SMLP-query &    yes &   yes &   yes &   no & yes &   yes &            no &     yes &         no &         no \\
    SMLP-verify &    yes &   yes &   yes &   no & yes &   yes &            no &      no &        yes &         no \\
SMLP-synthesize &    yes &   yes &   yes &  yes & yes &   yes &            no &      no &        yes &         no \\
  SMLP-optimize &    yes &   yes &   yes &  yes & yes &   yes &           yes &      no &         no &        yes \\
    SMLP-optsyn &    yes &   yes &   yes &  yes & yes &   yes &           yes &      no &        yes &        yes \\
\bottomrule
\end{tabular}
\caption{Mapping SMLP modes to relevant features}
\label{smlp:modes:features}
\end{table}

\subsection{Exploration basic concepts}\label{sub:exploration:concepts}

A formal definition of the tasks accomplished with these model exploration modes can be found in~\cite{brausse2024smlp}. 
Formal descriptions combined with informal clarifications will be provided in this section as well. 
Running SMLP in these modes reduces to solving formulas of the following structure:
\begin{equation}\label{form:gear:final}
    \exists p ~\big[ \eta(p) \wedge
    \forall p'~
    \forall x y~[
    \theta(p,p') \implies (\varphi_M(p',x,y)  \implies  \varphi_{\mathit{cond}}(p',x,y))
    ]\big]
\end{equation}
where $p$ denotes model knobs, $x$ denotes inputs, $y$ denotes the outputs, $\eta(p)$ defines 
constraints on the knobs, $\theta(p,p')$ defines the stability region, $\varphi_M(p',x,y)$ defines the ML model constraints, and the condition
$\varphi_{\mathit{cond}}(p',x,y)$ depends on the SMLP mode and will be discussed in subsections below.
$\eta(p)$ can be constraints on individual knobs or more complex constraints defining relationships between knobs.
The stability region $\theta(p,p')$ can be any reflexive predicate 
and in SMLP we use
$\theta_r(p, p') \eqdef \Vert p - p' \Vert \leq r$, where  $\Vert p - p' \Vert$ is a distance between two configurations $p$ and $p'$, and $r$ is a relative or absolute \emph{radius}; that is, 
the $\theta_r(p, p')$ region corresponds to a ball (or box) around $p$.
$\varphi_M(p,x,y)$ is computed by SMLP internally, based on the ML model specification.
$\varphi_{\mathit{cond}}(p',x,y)$  can represent (1) verification conditions, (2) model querying conditions, 
(3) parameter optimization conditions, or (4) parameter synthesis conditions.

\begin{defn}\label{def:stable:witness:validity}
\begin{itemize}
\item 
Given a value assignment $p^*$ to knobs $p$, an assignment $x^*$ to inputs $x$ is called a \emph{$\theta$-stable witness} 
to $\formula(p,x,y)$ for configuration $p^*$ if the following formula is valid:
\begin{equation}\label{form:gear:certify}
\varphi_{\mathit{certify}}^{px}(p^*,x^*) \eqdef
     \eta(p^*) \wedge ~\big(
    \forall p'~
    \forall y~[
   \theta(p^*, p') \implies (\varphi_M(p',x^*,y)  \implies  \varphi_{\mathit{cond}}(p',x^*,y))
    ]\big)
\end{equation}
where \[\varphi_{\mathit{cond}}(p,x,y) \eqdef \alpha(p,x) \implies \formula(p,x,y).\] 
Here, $\formula(p,x,y)$ can be either $\query,\assert$ or $\beta$, described below.

\item 
A value assignment $p^*$ to knobs $p$ is called a  \emph{$\theta$-stable configuration for $\formula(p,x,y)$} if any legal value assignment 
$x^*$ to inputs $x$ is a $\theta$-stable witness to $\formula(p,x,y)$ for configuration $p^*$; that is, when the following formula is valid:
\begin{equation}\label{form:gear:verify}
\varphi_{\mathit{verify}}^{p}(p^*) \eqdef
     \eta(p^*) \wedge ~\big(
    \forall p'~
    \forall xy~[
   \theta(p^*, p') \implies (\varphi_M(p',x,y)  \implies  \varphi_{\mathit{cond}}(p',x,y))
    ]\big)
\end{equation}
\item 
If $x^*$ is a  $\theta$-stable witness for $\formula(p, x,y)$ for configuration $p^*$, then we call $x^*$  a \emph{$\theta$-stable counter-example to  
assertion $\assert(p^*, x,y)  \eqdef \neg \formula(p^*, x,y)$, for configuration $p^*$}.
\delete{
An assertion $\assert(p, x,y)$ is called \emph{$\theta$-valid with respect to knob configuration $p^*$} if its negation 
$\query(p, x,y) \eqdef \neg \assert(p, x,y)$ does not have a $\theta$-stable witness $x^*$ for $p^*$.  In the latter case  
assertion $\assert(p^*, x,y)$ is called \emph{$\theta$-valid}. Otherwise, if $x^*$ is a  $\theta$-stable witness for 
$\query(p^*, x,y)$, then we call $x^*$  a \emph{$\theta$-stable counter-example to  $\assert(p^*, x,y)$}, and 
$\assert(p^*, x,y)$ is \emph{$\theta$-falsifiable} (that is, it is not $\theta$-valid).
}
\end{itemize}
\end{defn}

\Cref{form:gear:certify} corresponds to mode $\mode{certify}$ in SMLP, and  \Cref{form:gear:verify} 
corresponds to mode $\mode{verify}$, and they will be discussed in detail in Subsections~\ref{sect:certify} and \ref{sect:verify}, 
respectively. The concept of \emph{$\theta$-stable counter-example} to  an assertion is relevant for targeted model refinement
loop for verifying assertions, and is discussed in Section~\ref{sec:refinement}.

\ZK{it would be useful to define sufficient and necessary condition for vacuity of the formulas describing model exploration modes.}
The \emph{interface consistency check} is defined as simply checking satisfiability  of: 
\begin{equation}\label{form:interface:consistency}
\alpha(p, x) \wedge \eta(p). 
\end{equation}

The \emph{model consistency} check augments the interface consistency check with consistency of the constraints $\varphi_M(p,x,y)$ 
defining the ML model $M$ in conjunction with $\alpha$ and $\eta$ constraints: model consistency check is defined as checking satisfiability of:
\begin{equation}\label{form:model:consistency}
\alpha(p, x) \wedge \eta(p) \wedge \varphi_M(p,x,y). 
\end{equation}

These two consistency checks are common for all model exploration modes, because when one of these checks fails, 
then model exploration task is not well defined. 

\delete{
One other vacuity related check which is common to all model exploration modes is the \emph{configuration consistency check}.
It is defined as checking the satisfiability of:  

\begin{equation}\label{form:configuration:consistency}
\alpha(p, x) \wedge \eta(p) \wedge \vartheta(p,x)  \wedge \varphi_M(p,x,y) \wedge \beta(p,x,y),
\end{equation}
\noindent where $\vartheta(p,x)$ is the formula inferred form value assignments to $p,x$ in the witness for a query in 
$\mode{certify}$ mode, value assignments to $p$ in the configuration for an assertion in the $\mode{verify}$ mode, 
and is true in other modes; and $\beta(p,x,y)$ is true in $\mode{certify}$ and $\mode{verify}$ modes.
\ZK{In fact, $ \vartheta_{witn}(p,x)$ plays the role of $\beta(p,x,y)$ when the latter does not actually depend on $y$..
Better concept might be (1) to use $\vartheta_{conf}$ instead of $\vartheta_{witn}$, that is, evan in certification mode only 
knobs are assigned in the witness while inputs are not assigned, and (2) to not refer to  $\vartheta_{witn}$ in the formula and instead 
say $\beta$ is  $\vartheta_{conf}$ when the mode is certify or verify.}
}

\subsection{Mode $\mode{certify}$: certification of a stable witness}\label{sect:certify}

In the $\mode{certify}$ mode, SMLP is given an assignment $p^*,x^*$ to knobs $p$ and inputs $x$, 
 a query $\query(p,x,y)$ on an ML model $M$, and we want to check whether $x^*$ is a stable witness to
 $\query(p,x,y)$ for $p^*$, as defined in Definition~\ref{def:stable:witness:validity}:
certification requires checking validity of \cref{form:gear:certify}, with $\formula(p,x,y) \eqdef \query(p,x,y)$ in
$\varphi_{\mathit{cond}}(p,x,y)$. 
Currently SMLP assumes that all knobs in $p$ and all inputs in $x$ are assigned concrete values in $p^*, x^*$.
This requirement can be relaxed.

\begin{figure}
\center
\includegraphics[width= 0.5\columnwidth]{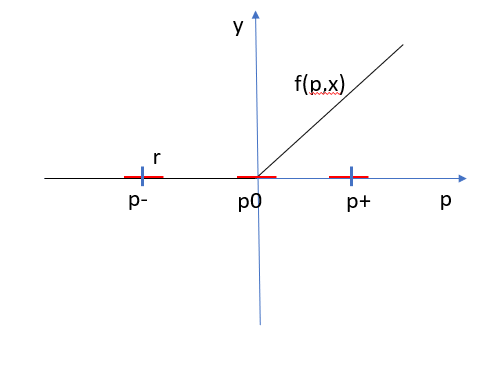}
\caption{Model exploration running example, $x$ is projected away to simplify the graph.}
\label{fig:certi:veri:query:synth}
\end{figure}

\begin{exmpl}\label{example:certify}
(Running Example) Let's assume that function $y = f(p,x)$ defined in \cref{func:model:exploration}, depicted in 
Figure~\ref{fig:certi:veri:query:synth}, represents a model with knob $p$, input $x$, 
and output $y$, with constraint $\alpha \eqdef -2 \leq p \leq 2 \,\wedge\, -1 \leq x \leq 1$. 
Let's consider three concrete values for $p$ within its range $[-2,2]$: $p^*_- <0$, 
$p^*_0 = 0$, and $p^*_+ > 0$, and two concrete values $x^*_- < 0$ and $x^*_+ > 0$ for $x$ in its range $[-1,1]$. 
And let $\query(p,x,y) \eqdef y \leq 0$. Then $x^*_-$ is a $\theta_r$-stable witness to 
$\query$ for $p^*_-$, for any $0 < r \leq \lvert p^*_- \rvert$;  $x^*_-$ is a $\theta_0$-stable witness to $\query$ 
for $p^*_0$ (meaning, $x^*_-$ satisfies $\query$ and is $\theta_r$-stable witness for 
$r = 0$), but is not a $\theta_r$-stable witness 
for any $r >0$ (because $\query$ evaluates to $\operator{false}$ for positive values of $p$,
and $\theta_r$-stability region of $p^*_0$ contains legal positive values of $p$ for any $r >0$); 
and $x^*_-$ is not a witness to $\query$ for $p^*_+$. Finally, $x^*_+$
is not a witness, and therefore not a stable witness, for any legal value of $p$
(because $\query$ evaluates to $\operator{false}$ for positive values of~$x$). 

\begin{equation}\label{func:model:exploration}
  \text{y = f(p,x)} =
  \begin{cases}
    0 & \text{$p \leq 0 \wedge x \leq 0$} \\
    \text{x} & \text{$x > 0$} \\
    \text{p} & \text{otherwise}
  \end{cases}
\end{equation}
\end{exmpl}

SMLP supports certification of multiple queries in a single run, and each query is certified with respect to its corresponding
witness (different queries might refer to different witnesses). The general (and the recommended) way of
defining a witness per query is using the $\speckey{witnesses}$ field in SMLP spec file. The  $\speckey{witnesses}$
field is a dictionary with query names as keys and the respective values are dictionaries assigning a concrete value to 
each knob and each input. An example is displayed in Figure~\ref{fig:certify:witnesses:spec} where say  
$\speckey{query\_stable\_witness}$ is the name of a query, $\speckey{p1}$ and $\speckey{p2}$ are names of knobs, 
and $\speckey{x}$ is the name of the input.

\begin{figure}[tp]
\small
\begin{verbatim}
        "witnesses": {
                "query_stable_witness": {
                        "x": 7,
                        "p1": 7.0,
                        "p2": 6.000000067055225
                },
                "query_grid_conflict": {
                        "x": 6.2,
                        "p1": 3.0,
                        "p2": 6.000000067055225
                },
                "query_unstable_witness": {
                        "x": 7,
                        "p1": 7.0,
                        "p2": 6.0
                },
                "query_infeasible_witness": {
                        "x": 7,
                        "p1": 7.0,
                        "p2": 6.0
                }
        }
\end{verbatim}
\caption{SMLP example of specifying witnesses in mode $\mode{certify}$.}
\label{fig:certify:witnesses:spec}
\end{figure}

When the same witness is certified against all queries, and witnesses per query are not defined using the  $\speckey{witnesses}$ field,
SMLP applies a sanity check to see whether unique values $p^*$ to knobs $p$ and unique values $x^*$ to inputs $x$ can be inferred 
from knob and input ranges and knob grids specified in the spec file using the $\speckey{variables}$ field. 
When values $p^*, x^*$ cannot be inferred this way, SMLP aborts certification with a message clarifying the reason.

Below $\vartheta_{p^*x^*}(p,x)$ denotes the formula inferred from value 
assignments to $p,x$ in a given witness to a query when the witness is specified using the
$\speckey{witnesses}$ field in the spec file, and is constant true otherwise. 
For example, if $p$ and $x$ are single variable vectors,
$p^* = 3$ and $x^* = 2$, then 
$\vartheta_{p^*x^*}(p,x) \eqdef (p = 3 \wedge x = 2)$. (In the spec file we use Python \texttt{==} in place of $=$.)

As discussed in Section~\ref{sub:exploration:concepts}, the interface consistency \cref{form:interface:consistency} and
model consistency \cref{form:model:consistency} 
checks are performed before starting actual certification of witnesses for stability with respect to the corresponding queries. 
In addition to these, for the certification problem in \cref{form:gear:certify} to be well defined, 
we need to check that $\eta(p^*) \wedge \alpha(p^*,x^*)$ evaluates to constant true: 
this means that $p^*, x^*$ witnesses that $\alpha(p,x) \wedge \eta(p)$ is \emph{consistent}.
To re-iterate, the \emph{witness consistency check} for the certification task consists in checking satisfiability of:
\begin{equation}\label{form:gear:certify:witness:consistency}
   \vartheta_{p^*x^*}(p,x) \wedge  \alpha(p, x) \wedge \eta(p)
\end{equation}

In the implementation, we are using a stronger version of the witness consistency check  \cref{form:gear:certify:witness:consistency}, 
which in addition takes into account the ML model constraints, and consists in checking satisfiability of:
\begin{equation}\label{form:certify:witness:model:consistency}
\vartheta_{p^*x^*}(p,x) \wedge  \alpha(p, x) \wedge \eta(p) \wedge \varphi_M(p,x,y)
\end{equation}

Note that none of the above consistency checks refers to an actual query $\query(p,x,y)$ for which certification is performed.
The \emph{feasibility} part for this task is checking satisfiability of 

\begin{equation}\label{form:gear:certify:feasibility}
      \vartheta_{p^*x^*}(p,x) \wedge  \alpha(p, x) \wedge \eta(p)  \wedge  \varphi_M(p,x,y) \wedge  \query(p,x,y))
\end{equation}

If the above formula is not satisfiable, then $x^*$ cannot be a stable witness to $\query(p,x,y)$ for $p^*$. 
Otherwise stability of the candidate witness $p^*, x^*$ to 
$\query(p,x,y)$ is checked by proving validity of \cref{form:gear:certify},
and this is done by checking satisfiability of \cref{form:gear:certify:feasibility:sat}:

\begin{equation}\label{form:gear:certify:feasibility:sat}
       \vartheta_{p^*x^*}(p,x) \wedge \eta(p) \wedge \theta(p,p') \wedge \alpha(p', x)   \wedge  \varphi_M(p',x,y) \wedge \neg \query(p',x,y))
\end{equation}

An example command to run SMLP in $\mode{certify}$ mode is given in \Cref{fig:command:certify}.

\begin{figure}[hb!] 
\begin{verbatim}
../../src/run_smlp.py -data ../data/smlp_toy_ctg_num_resp -out_dir ./ -pref Test128 \
-mode certify -resp y1,y2 -feat x,p1,p2 -model poly_sklearn -dt_sklearn_max_depth 15 \
-save_model f -use_model f -model_per_response f -quer_names \
query_stable_witness,query_grid_conflict,query_unstable_witness,query_infeasible_witness\
-quer_exprs "y2<=90;y1>=9;y1>=(-10);y1>9" -plots f -seed 10 -log_time f  \
-spec ../specs/smlp_toy_witness_certify.spec
\end{verbatim}
\caption{Example of SMLP's command in mode $\mode{certify}$.}
\label{fig:command:certify}
\end{figure}

Certification results are reported in file $\suffix{prefix\_dataname\_certify\_results.json}$.
Figure~\ref{fig:certify:result} displays an example results file in $\mode{certify}$ mode.

\begin{enumerate}
\item Fields $\speckey{smlp\_execution}$,  $\speckey{interface\_consistent}$ and $\speckey{model\_consistent}$
are common for each (witness, query) pair, and they provide status of the entire execution of SMLP.
\item The field  $\speckey{witness\_consistent}$ specifies result of witness consistency check \cref{form:certify:witness:model:consistency}
\item The field  $\speckey{witness\_feasible}$ specifies result of witness feasibility check \cref{form:gear:certify:feasibility}
\item The field  $\speckey{witness\_status}$ specifies the witness certification status, and can be one of:
\begin{itemize}
\item[$\specval{''ERROR''}$]  when interface consistency \cref{form:interface:consistency},
model consistency \cref{form:model:consistency}, or witness consistency \cref{form:certify:witness:model:consistency} check fails.
\item[$\specval{''PASS''}$]  when interface consistency \cref{form:interface:consistency}, 
model consistency \cref{form:model:consistency},  configuration consistency \cref{form:certify:witness:model:consistency} 
and  witness validity \cref{form:gear:certify}  are all valid.
\item[$\specval{"FAIL"}$]  when interface consistency \cref{form:interface:consistency}, 
model consistency \cref{form:model:consistency},  witness consistency \cref{form:certify:witness:model:consistency} 
are all valid and  witness validity \cref{form:gear:certify}  is not valid. 
Query feasibility \cref{form:gear:certify:feasibility} may or may not be satisfiable, and this info is reported
using field $\speckey{witness\_feasible}$.
\item[$\specval{"UNKNOWN"}$] otherwise. This can happen when SMLP run terminates before one of the above 
results can be concluded.
\end{itemize}
\end{enumerate}

\begin{figure}[tp]
\small
\begin{verbatim}
{
        "query_stable_witness": {
                "witness_consistent": "true",
                "witness_feasible": "true",
                "witness_stable": "true",
                "witness_status": "PASS"
        },
        "query_grid_conflict": {
                "witness_consistent": "false",
                "witness_feasible": "false",
                "witness_stable": "false",
                "witness_status": "ERROR"
        },
        "query_unstable_witness": {
                "witness_consistent": "true",
                "witness_feasible": "true",
                "witness_stable": "false",
                "witness_status": "FAIL"
        },
        "query_infeasible_witness": {
                "witness_consistent": "true",
                "witness_feasible": "false",
                "witness_stable": "false",
                "witness_status": "FAIL"
        },
        "smlp_execution": "completed",
        "interface_consistent": "true",
        "model_consistent": "true"
}
\end{verbatim}
\caption{SMLP result in mode $\mode{certify}$.}
\label{fig:certify:result}
\end{figure}

\subsection{Mode $\mode{query}$: querying for a stable witness}

The task of querying ML model for a stable witness to query $\query(p, x, y)$ consists in finding value assignments  
$p^*,x^*$  for knobs $p$ and inputs $x$ that represent a solution for \cref{form:gear:synthesis:query:exists}:

\begin{equation}\label{form:gear:synthesis:query:exists}
    \exists p, x ~\big[ \eta(p) \wedge
    \forall p'~
    \forall y~[
    \theta(p,p') \implies (\varphi_M(p',x,y)  \implies  \varphi_{\mathit{cond}}(p',x,y))
    ]\big]
\end{equation} 
where \[\varphi_{\mathit{cond}}(p,x,y) \eqdef \alpha(p,x) \implies \query(p,x,y).\]
According to Definition~\ref{def:stable:witness:validity}, for any solution $p^*,x^*$ of \cref{form:gear:synthesis:query:exists},
$x^*$ is a stable witness for  $\query(p, x, y)$, for configuration $p^*$.

\begin{exmpl}\label{example:query}
(Running Example~\ref{example:certify}, continued) Consider again the model given by function $f(p,x)$ 
in \cref{func:model:exploration},  Figure~\ref{fig:certi:veri:query:synth}. For any $p^* \leq -r$, any 
value $-1 \leq x^*\leq 0$ is a $\theta_r$-stable witness to $\query \eqdef y \leq 0$ for $p^*$. Hence  
querying the model $f(p,x)$ will be successful in SMLP for any stability radius $0\leq r \leq 2$ 
(recall that the domain of $p$ is $[-2, 2]$) and any of the above pairs of values of $p^*, x^*$,
 and only those, can be returned by SMLP as a solution to querying the model $f(p,x)$ for condition $\query$.
\end{exmpl}

As already stated in Section~\ref{sub:exploration:concepts}, the interface consistency \cref{form:interface:consistency} and
model consistency \cref{form:model:consistency} checks are performed before starting actual querying for a stable witnesses.
This ensures that querying is well defined, and cannot fail  vacuously in case one of the above checks fail.

\ZK{Check these two formulas or get rid of them.} First, we find a candidate $p^*, x^*$ by solving
\begin{equation}\label{form:gear:synthesis:query}
    \exists p, x, y ~\big[ \eta(p) \wedge
    \varphi_M(p,x,y)  \wedge  \varphi_{\mathit{cond}}(p,x,y)
    ]\big]
\end{equation} 

If such $p^*, x^*$  exist, SMLP reports this in the results file using field  $\speckey{query\_feasible}$, by setting it to $\speckey{true}$; otherwise the query fails. If such $p^*, x^*$ exist, SMLP checks 
whether the following formula is valid (by checking its negation for satisfiability): 
\begin{equation}\label{form:gear:synthesis:query'}
    \eta(p^*) \wedge
    \forall p'~
    \forall y~[
    \theta(p^*,p') \implies (\varphi_M(p',x^*,y)  \implies  \varphi_{\mathit{cond}}(p',x^*,y))
    ]
\end{equation} 

If the above formula is valid, then we have shown that $x^*$ is a stable witness for  $\query(p^*, x, y)$, 
and the task is accomplished -- SMLP reports $p^*, x^*$ is solution to the synthesis task. Otherwise,
search should continue by searching for a solution different from $p^*, x^*$ (and their neighborhood). \KK{This is a reduced version of the GearSAT alg maybe refer to it?}

An example command to run SMLP in $\mode{query}$ mode is given in \Cref{fig:command:query}.

\begin{figure}[hb!] 
\begin{verbatim}
../../src/run_smlp.py -data "../data/smlp_toy_basic" -out_dir ./ -pref Test119 \
-mode query -model system -resp y1,y2 -feat p1,p2 -save_model f -use_model f \
-mrmr_pred 0 -model_per_response t -plots f -seed 10 -log_time f \
-spec ../specs/smlp_toy_system_stable_constant_query.spec
\end{verbatim}
\caption{Example of SMLP's command in mode $\mode{query}$.}
\label{fig:command:query}
\end{figure}

Query results are reported in file $\suffix{prefix\_dataname\_query\_results.json}$.
SMLP supports querying multiple conditions in one SMLP run. 
Figure~\ref{fig:query:result} displays an example results file in $\mode{query}$ mode,
where the  $\speckey{query\_result}$ field 
specifies the values of knobs, as well as values of inputs and values of outputs
found in the satisfying assignment to~\cref{form:gear:certify:feasibility} that identified the stable witness.
\begin{figure}[tp]
\small
\begin{verbatim}
{
        "smlp_execution": "completed",
        "interface_consistent": "true",
        "query_feasible_unstable": {
                "query_feasible": "true",
                "query_stable": "false",
                "query_status": "FAIL",
                "query_result": null
        },
        "query_feasible_stable": {
                "query_feasible": "true",
                "query_stable": "true",
                "query_status": "PASS",
                "query_result": {
                        "p1": 0.0,
                        "y2": 0.0,
                        "p2": 0.0,
                        "y1": 0.0
                }
        },
        "query_infeasible": {
                "query_feasible": "false",
                "query_stable": "false",
                "query_status": "FAIL",
                "query_result": null
        },
        "model_consistent": "true"
}
\end{verbatim}
\caption{SMLP result in mode $\mode{query}$.}
\label{fig:query:result}
\end{figure}

\subsection{Mode $\mode{verify}$: assertion verification with stability}\label{sect:verify}

In assertion verification usually one assumes that the knobs have already been fixed to legal values, 
and their impact has been propagated through the constraints, therefore usually in the context of assertion 
verification knobs are not considered explicitly. However, in order to formalize stability also in the context of 
verification, SMLP assumes that the values of knobs $p$ are assigned constant values $p^*$, but can be 
perturbed (by environmental effects or by an adversary), therefore treatment of $p^*$ is explicit. Then the problem of 
verifying an assertion $\assert(p,x,y)$ for configuration $p^*$ under allowed perturbations $p'$ of knob values $p^*$ 
controlled by $\theta(p^*, p')$, is exactly the problem of checking stability of configuration $p^*$ for assertion 
$\assert(p,x,y)$, as defined in Definition~\ref{def:stable:witness:validity}. To re-iterate, the problem of 
\emph{verification with stability} is formalized in SMLP as validity of~\cref{form:gear:verify}.

\begin{exmpl}\label{example:verify}
(Running Example~\ref{example:certify}, continued)
For the model given by function $y = f(p,x)$ in \cref{func:model:exploration},  
Figure~\ref{fig:certi:veri:query:synth}, let us define $\assert \eqdef y \leq 0$. 
Then for any legal vale $p^*$ of $p$, $\assert$ fails for positive values of $x$ 
(even if the stability radius $r = 0$). However, if the range of $x$ is restricted 
$-1 \leq x \leq 0$, then for any knob configuration  $p^* \leq 0$, $\assert$ passes verification 
with stability for any radius $r \leq \lvert p^* \rvert$. Note that for configuration $p^* = 0$,
while $\assert$ is valid without stability requirements for the restricted range of $x$, $\assert$ fails 
verification with stability for any radius~$r>0$ (because $y = f(p,x) > 0$ for positive values of~$p$).
\end{exmpl}

SMLP supports verification of multiple assertions in a single run, and each assertion is verified with respect to its corresponding
configuration (different assertions might refer to different configurations). The general (and the recommended) way of
defining a configuration per assertion is using the $\speckey{configuration}$ field in SMLP spec file. The  $\speckey{configuration}$
field is a dictionary with assertion names as keys and the respective values are dictionaries assigning a concrete value to each knob.
An example is displayed in Figure~\ref{fig:verify:witness:spec}, where say  $\speckey{stable\_configuration}$ is the name of an assertion
and $\speckey{p1}$ and $\speckey{p2}$ are names of knobs.

\begin{figure}[tp]
\small
\begin{verbatim}
       "configurations": {
                "stable_config": {
                        "p1": 7.0,
                        "p2": 6.000000067055225
                },
                "grid_conflict": {
                        "p1": 3.0,
                        "p2": 6.000000067055225
                },
                "unstable_config": {
                        "p1": 7.0,
                        "p2": 6.0
                },
                "not_feasible": {
                        "p1": 7.0,
                        "p2": 6.0
                }
        }
\end{verbatim}
\caption{SMLP specification example to specify $\speckey{configurations}$ per assertion, in mode $\mode{verify}$.}
\label{fig:verify:witness:spec}
\end{figure}

When each assertion is verified with respect to the same configuration, and configurations per assertions are not defined using 
the  $\speckey{configurations}$ field, SMLP applies a sanity check to see whether unique values $p^*$ to knobs $p$ can be inferred 
from knob ranges and grids specified in the spec file using the $\speckey{variables}$ field. When values $p^*$ cannot be inferred 
this way, SMLP aborts verification with an error message clarifying the reason.

Below $\vartheta_{p^*}(p)$ denotes the formula inferred form value assignments to $p$ in 
the configuration for the corresponding assertion when the latter is specified in the spec file using field
$\speckey{configurations}$, and is constant true otherwise. For example, if $p$ is a variable vector $p1,p2$,  
then $\vartheta_{p^*}(p)$ might look like  $\vartheta_{p^*}(p1,p2)\eqdef (p1 = 3 \wedge p2 = 5)$.

As discussed in Section~\ref{sub:exploration:concepts}, the interface consistency \cref{form:interface:consistency} and
model consistency \cref{form:model:consistency} 
checks are performed before starting actual assertion verification. In addition to these,
for the verification problem \cref{form:gear:verify} to be well defined, we need to check that $\eta(p^*)$ evaluates to constant true 
and  $\alpha(p^*,x)$ is satisfiable. This means that $p^*$ witnesses that $\alpha(p,x) \wedge \eta(p)$ is \emph{consistent}, 
that is, there exist values of inputs that satisfy  $\alpha(p^*,x) \wedge \eta(p^*)$.  Formally, \emph{configuration consistency} 
check for verification requires, as a necessary condition (but not a sufficient condition), the following formula to be valid: 

\begin{equation}\label{form:gear:verify:consistency:interface}
\exists x~ [\vartheta_{p^*}(p) \wedge \alpha(p, x) \wedge \eta(p)]
\end{equation}

In the implementation, we are using a stronger version of the configuration interface consistency check, 
called \emph{configuration consistency} check, which in 
addition takes into account the ML model constraints when checking consistency of the witness;
this check subsumes satisfiability check for \cref{form:gear:verify:consistency:interface}:
\begin{equation}\label{form:verify:witness:model:consistency}
\exists x~ [\vartheta_{p^*}(p) \wedge \alpha(p, x) \wedge \eta(p)  \wedge \varphi_M(p,x,y)]
\end{equation}

Before performing verification according to \cref{form:gear:verify}, SMLP checks satisfiability 
of the following formula, which we refer to as \emph{assertion feasibility} part of verification with stability:
\begin{equation}\label{form:gear:verify:feasibility}
\exists x~ [\vartheta_{p^*}(p) \wedge \alpha(p, x) \wedge \eta(p)  \wedge \varphi_M(p,x,y) \wedge  \assert(p,x,y)]
\end{equation}

\noindent If the above formula is not satisfiable, then the negated assertion will be true for any legal inputs, which means that the assertion 
fails everywhere in the legal input space. This is useful info because in such a case it can be that the components of the problem instance 
(for example, the assertion or the constraints) were not specified correctly.

Next, stability of the configuration $p^*$ for $\assert(p,x,y)$ is checked by proving validity of formula \cref{form:gear:verify},
and this is done by checking satisfiability of formula~\cref{form:gear:verify:negated}, which is the negation of \cref{form:gear:verify}:
\begin{equation}\label{form:gear:verify:negated}
\vartheta_{p^*}(p) \wedge \theta(p, p')  \wedge  \varphi_M(p',x,y)  \wedge  \alpha(p',x) \wedge   \neg \assert(p',x,y))
\end{equation}

An example command for mode $\mode{verify}$ is given in \Cref{fig:command:verify}
\begin{figure}[hb!] 
\begin{verbatim}
../../src/run_smlp.py -data ../data/smlp_toy_ctg_num_resp -out_dir ./ \
-pref Test129 -mode verify -resp y1,y2 -feat x,p1,p2 -model poly_sklearn \
-save_model f -use_model f -model_per_response f -asrt_names \
assert_stable_config,assert_grid_conflict,assert_unstable_config,assert_infeasible \
-asrt_exprs "y2<=90;y1>=9;y1>=(-10);y1>20" -plots f -seed 10 -log_time f  \
-spec ../specs/smlp_toy_configuration_verify.spec
\end{verbatim}
\caption{Example of SMLP's command in mode $\mode{verify}$.}
\label{fig:command:verify}
\end{figure}

Verification results are reported in file $\suffix{prefix\_dataname\_verify\_results.json}$.
Figure~\ref{fig:verify:result} displays an example results file of SMLP in $\mode{verify}$ mode.

\begin{enumerate}
\item Fields $\speckey{smlp\_execution}$,  $\speckey{interface\_consistent}$ and $\speckey{model\_consistent}$
are common for each assertion, and they provide status of the entire execution of SMLP.
\item The field  $\speckey{configuration\_consistent}$ specifies result of \cref{form:verify:witness:model:consistency}
\item The field  $\speckey{assertion\_feasible}$ specifies result of \cref{form:gear:verify:feasibility}
\item The field  $\speckey{assertion\_status}$ specifies the assertion verification status, and can be one of:
\begin{itemize}
\item[$\specval{''ERROR''}$]  when interface consistency \cref{form:interface:consistency} or
model consistency \cref{form:model:consistency} or configuration consistency \cref{form:verify:witness:model:consistency} s not valid.
\item[$\specval{''PASS''}$]  when interface consistency \cref{form:interface:consistency}, 
model consistency \cref{form:model:consistency},  configuration consistency \cref{form:verify:witness:model:consistency} 
and  assertion validity \cref{form:gear:verify}  are all valid.
\item[$\specval{"FAIL"}$]  when interface consistency \cref{form:interface:consistency}, 
model consistency \cref{form:model:consistency},  configuration consistency \cref{form:verify:witness:model:consistency} 
are all valid and  assertion validity \cref{form:gear:verify}  is not valid.
Assertion feasibility \cref{form:gear:verify:feasibility} may or may not be satisfiable, and this is reported using
field $\speckey{assertion\_feasible}$.
\item[$\specval{"UNKNOWN"}$] otherwise.  This can happen when SMLP run terminates before one of the above 
results can be concluded.
\end{itemize}
\end{enumerate}

\begin{figure}[tp]
\small
\begin{verbatim}
{
        "assert_stable_config": {
                "configuration_consistent": "true",
                "assertion_status": "PASS",
                "counter_example": null,
                "assertion_feasible": true
        },
        "assert_grid_conflict": {
                "configuration_consistent": "false",
                "assertion_status": "ERROR",
                "counter_example": null,
                "assertion_feasible": "false"
        },
        "assert_unstable_config": {
                "configuration_consistent": "true",
                "assertion_status": "FAIL",
                "counter_example": {
                        "y2": 55.69463654220261,
                        "y1": -58.38640996591811,
                        "p2": 6.125,
                        "p1": 7.5,
                        "x": 1.0
                },
                "assertion_feasible": true
        },
        "assert_infeasible": {
                "configuration_consistent": "true",
                "assertion_status": "FAIL",
                "counter_example": {
                        "y2": 55.69463654220261,
                        "y1": -58.38640996591811,
                        "p2": 6.125,
                        "p1": 7.5,
                        "x": 1.0
                },
                "assertion_feasible": false
        },
        "smlp_execution": "completed",
        "interface_consistent": "true",
        "model_consistent": "true"
}
\end{verbatim}
\caption{SMLP result in mode $\mode{verify}$.}
\label{fig:verify:result}
\end{figure}

\subsection{Mode $\mode{synthesize}$: parameter synthesis with stability}

The task of $\theta$-stable synthesis consists of finding a solution to formula~\cref{form:gear:final}, where
\[\varphi_{\mathit{cond}}(p,x,y) \eqdef \alpha(p,x) \implies (\beta(p, x, y) \wedge \assert(p,x,y))\]
and  $\assert(p, x, y)$ might represent a conjunction of multiple assertions.
According to Definition~\ref{def:stable:witness:validity}, any solution $p^*$ to the synthesis problem,  
$p^*$  is a stable witness for $\formula(p, x, y) \eqdef \beta(p, x, y) \wedge \assert(p,x,y)$; the latter
expresses the constraints that synthesized design should satisfy (in legal input space).

\begin{exmpl}\label{example:synthesize}
(Running Example~\ref{example:certify}, continued)
For the model given by function $y = f(p,x)$ in \cref{func:model:exploration},  
Figure~\ref{fig:certi:veri:query:synth}, let us define $\beta \eqdef y \leq 0$
and $\assert = \operator{true}$. Then, for any legal value $p^*$ of $p$, $\beta$ evaluates 
to $\operator{false}$ for positive values of $x$;
therefore synthesis that guarantees validity of $\beta$ is not feasible 
(even for the stability radius $r = 0$). However, if the range of $x$ is restricted 
to $-1 \leq x \leq 0$, then for any knob configuration  $p^* \leq 0$, 
$\beta$ is valid (for any values of $x$ in its restricted range)
for any stability radius $r \leq \lvert p^* \rvert$, hence stable synthesis is feasible. 
Note that for configuration $p^* = 0$, $\beta$ is valid for stability radius $r = 0$,
therefore the usual synthesis procedure (that does not take stability requirements 
into account) might synthesize the model $f(p,x)$ into configuration $f(0,x)$, 
which will not be robust against perturbations or inaccuracies in measurements or modeling,
thus will not be reliable for exploring the system modeled by $f(p,x)$. 
\end{exmpl}

Just like for any other mode of exploration, the interface consistency \cref{form:interface:consistency} and
model consistency \cref{form:model:consistency} checks are performed before starting actual synthesis procedure.
This ensures that synthesis task is well defined, and cannot fail vacuously in case one of the above checks fail.

An example command for mode $\mode{synthesize}$ is given in \Cref{fig:command:synthesize}. 
In this command, as the ML model we actually use a python expression specified in the spec file
using field $\speckey{system}$. This is an initial support for specifying systems in SMLP and will be 
developed and changed in the future, therefore we do not provide any further details here. Please note that support of
python expressions as ML models allows one to perform exploration of python expressions with SMLP, in all model exploration 
modes (e.g., querying, verification, optimization) by taking stability requirements into consideration.
\begin{figure}[hb!] 
\begin{verbatim}
../../src/run_smlp.py -data ../data/smlp_toy_basic -out_dir ./ -pref Test121 \
-mode synthesize -model system -resp y1,y2 -feat p1,p2 -save_model f \
-use_model f -mrmr_pred 0 -model_per_response t -plots f -seed 10 -log_time f \
-spec ../specs/smlp_toy_system_stable_constant_synth_feasible.spec
\end{verbatim}
\caption{Example of SMLP's command in mode $\mode{synthesize}$.}
\label{fig:command:synthesize}
\end{figure}

Synthesis result is reported in file $\suffix{prefix\_dataname\_synthesize\_results.json}$.
Figure~\ref{fig:synthesize:result} displays an example results file in $\mode{synthesize}$ mode.
The field $\speckey{configuration\_feasible}$ reports the validity of 
\begin{equation}\label{form:gear:query:candidate}
\varphi_{\mathit{feasible}} \eqdef  \exists p,x~ \big[ \eta(p) \wedge \big(
    \forall y~[
   (\varphi_M(p,x,y)  \implies  \varphi_{\mathit{cond}}(p,x,y))
    ]\big)\big]
\end{equation}
The field  $\speckey{synthesis\_result}$ reports the synthesis results (the synthesized configuration of knob values).
The field  $\speckey{configuration\_stable}$ reports that a stable configuration satisfying the synthesis requirements
have been found, which is the same as reporting $\speckey{synthesis\_status}$ as  $\specval{PASS}$.

\begin{figure}[tp]
\small
\begin{verbatim}
{
        "smlp_execution": "completed",
        "interface_consistent": "true",
        "model_consistent": "true",
        "configuration_feasible": "true",
        "configuration_stable": "true",
        "synthesis_status": "PASS",
        "synthesis_result": {
                "p1": 0.0,
                "p2": 0.0
        }
}
\end{verbatim}
\caption{SMLP result in mode $\mode{synthesize}$.}
\label{fig:synthesize:result}
\end{figure}

\subsection{Mode $\mode{optimize}$: multi-objective optimization with stability}\label{sec:stable:opt}

\begin{figure}[tp]
\center
\includegraphics[width= 0.7\columnwidth]{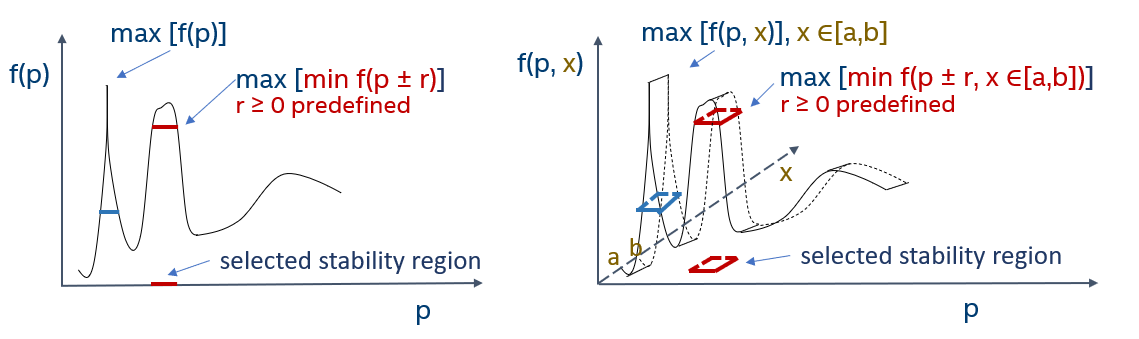}
\caption{SMLP max-min optimization. On both plots, $p$ denote the knobs. On the right plot we also consider inputs $x$ 
(which are universally quantified) as part of $f$.} \label{smlp_maxmin}
\end{figure}

In this subsection we consider the optimization problem for a real-valued function $f$ (in our case, an ML model),
extended in two ways:
\begin{enumerate}
\item We consider a $\theta$-stable maximum to ensure that the objective function  does not drop drastically in a 
close neighborhood of the configuration where its maximum is achieved.
\item We assume that the objective function besides knobs depends also on inputs, and the 
function is maximized in the stability $\theta$-region of knobs, for any values of inputs in their respective legal ranges.
\end{enumerate}
We explain these extensions using two plots in Figure~\ref{smlp_maxmin}.
The left plot represents optimization problem for $f(p,x)$ when $f$ depends on knobs only (thus $x$ is an empty vector), 
while the right plot represents the general setting where $x$ is not empty (which is usually not considered in optimization research). 
In each plot, the blue threshold (in the form of a horizontal bar or a rectangle) denotes the stable maximum around the point 
where $f$ reaches its (regular) maximum, and the red threshold denotes the stable maximum, which is approximated by our 
optimization algorithms. In both plots, the regular maximum of $f$ is not stable due to a sharp drop of $f$'s value in the stability region. 
\delete{The right plot depicts the situation when input $x$ ranges in interval $[a, b]$, thus in \emph{max-min} optimization problem, 
formalized below in Formulas~\eqref{form:opt1}  and  \eqref{form:opt2}, the minimum of $f$ is calculated in the stability region 
of knobs $p$, with values of $x$ ranging in $[a,b]$.}

An example of how to run SMLP in mode $\mode{optimize}$ was given in \Cref{sec:example}, which among other
things describes how the objectives can be defined through the command line and through the specification file.
When there are multiple objectives, SMLP supports both Pareto optimization as well as optimizing for each objective
separately (independently from requirements of other objectives). This choice is controlled using option 
$\optionval{-pareto}{t/f}$.

Just like for other model exploration modes, the interface consistency \cref{form:interface:consistency} and
model consistency \cref{form:model:consistency} checks are performed before starting actual optimization procedure.
If these checks are successful, SMLP optimization algorithm performs feasibility check that $\beta$ constraints are 
feasible under the interface constraints $\alpha$ and $\eta$, and if a solution is found, the input and knob values in the satisfying assignment demonstrating the
feasible configuration, along with the values of the responses and the objectives, 
are (immediately) reported to optimization report file  with suffix
$\suffix{*\_optimization\_progress.csv}$.

SMLP then continues search to tighten the objective's upper and lower bounds,
and at anytime when lower bounds are improved (in case of maximization problem) the optimization progress report
is updated with improved estimates of the optima. If the search terminates under given time and memory
requirements, the final results are reported in file with suffix  $\suffix{*\_optimization\_results.csv}$. Reports
$\suffix{*\_optimization\_progress.json}$ and  $\suffix{*\_optimization\_results.json}$ are also available with 
more detail compared to the respective $\suffix{*.csv}$ reports. 

The stable optimization problem is a special case of stable optimized synthesis problem which is discussed formally
in \Cref{sec:stable:opt:synthesis}, and we refer the reader to it  for a formal treatment of stable optimization problem. 
More precisely, the problem of stable optimization is defined using \Cref{form:opt2} and the 
problem of stable optimized synthesis problem is defined using \Cref{form:optsyn}, from which \Cref{form:opt2}
is obtained as a special case, by assuming that $\assert(p,x,y)$ is constant $\operator{true}$.

An example command for mode $\mode{optimize}$ is given in \Cref{fig:command:optimize}.
\begin{figure}[hb!] 
\begin{verbatim}
../../src/run_smlp.py -data "../data/smlp_toy_basic" -out_dir ./ -pref Test123 \
-mode optimize -pareto t -model system -resp y1,y2 -feat p1,p2 -save_model f \
-use_model f -mrmr_pred 0 -model_per_response t -epsilon 0.00000001 -plots f -seed 10 \
-log_time f -spec ../specs/smlp_toy_system_stable_constant_synth_feasible.spec
\end{verbatim}
\caption{Example of SMLP's command in mode $\mode{optimize}$.}
\label{fig:command:optimize}
\end{figure}

Figure~\ref{fig:optimize:result} displays an example results file in $\mode{optimize}$ mode.
\begin{figure}[tp]
\small
\begin{verbatim}
{
        "objv1": {
                "value_in_config": 0.0,
                "threshold_scaled": -0.022943015285783935,
                "threshold": 0.0,
                "max_in_data": 10.7007,
                "min_in_data": 0.24
        },
        "objv2": {
                "value_in_config": 0.0,
                "threshold_scaled": -0.010615194948015235,
                "threshold": 0.0,
                "max_in_data": 102.36396627,
                "min_in_data": 1.0752000000000002
        },
        "y2": {
                "value_in_config": 0.0,
                "value_in_system": 0
        },
        "p1": {
                "value_in_config": 0.0
        },
        "y1": {
                "value_in_config": 0.0,
                "value_in_system": 0
        },
        "p2": {
                "value_in_config": 0.0
        },
        "objv2_scaled": {
                "value_in_config": -0.010615194948015235
        },
        "threshold_lo_scaled": {
                "value_in_config": -0.010615194948015235
        },
        "threshold_lo": {
                "value_in_config": -0.010615194948015235
        },
        "threshold_up_scaled": {
                "value_in_config": -0.006959520828193561
        },
        "threshold_up": {
                "value_in_config": -0.006959520828193561
        },
        "max_in_data": {
                "value_in_config": 1.0
        },
        "min_in_data": {
                "value_in_config": 0.0
        },
        "smlp_execution": "completed",
        "interface_consistent": "true",
        "model_consistent": "true",
        "synthesis_feasible": "true"
}
\end{verbatim}
\caption{SMLP result in mode $\mode{optimize}$.}
\label{fig:optimize:result}
\end{figure}

\subsection{Mode $\mode{optsyn}$: optimized synthesis with stability}\label{sec:stable:opt:synthesis}

Let us first consider optimization without stability or inputs, i.e., far low corner in the exploration cube Figure~\ref{fig:cube}. 
Given a formula
$\varphi_M$ encoding the model, and an objective function 
$\objv:
\mathcal{D}_{\mathit{par}}\times
\mathcal{D}_\mathit{out}\to\mathbb R$, the
standard optimization problem solved by SMLP is stated by Formula~\eqref{form:opt_p}.
 \begin{equation}\label{form:opt_p}
 \regmax{\varphi_M}{\objv}
 \eqdef\mathop{\max\limits_{p}} \{z \mid  
    \forall y\changedto[{~(} \varphi_M(p,y)  \implies  
    \objv(p,y) \geq z
    \changedto])\}
\end{equation}
A solution to this optimization problem  is the pair $(p^*,\regmax{\varphi_M}{\objv})$, where 
$p^*{}\in\mathcal{D}_{\mathit{par}}$ is a value of parameters $p$ on which the maximum 
$\regmax{\varphi_M}{\objv}\in\mathbb R$ of the objective function $\objv$ is achieved
for the output $y$ of the model on $p^*$.
In most cases it is not feasible to exactly compute the maximum.
To deal with this, SMLP computes maximum with a specified accuracy. 
Consider $\varepsilon >0$.  We refer to  values $(\tilde p,\tilde z)$ as a solution to the optimization problem with
\emph{accuracy} $\varepsilon$, or \emph{$\varepsilon$-solution},
if $\tilde z\leq \regmax{ \varphi_M}{\objv}  <\tilde z+\varepsilon$ holds and
$\tilde z$ is a lower bound on the objective, i.e., $\forall y [ \varphi_M(p,y)  \implies  
    \objv(p,y) \geq \tilde z] $ holds.

Now, we consider \emph{stable optimized synthesis}, i.e., the top right corner of the exploration cube.
The problem can be formulated as the following Formula~\eqref{form:opt1}, expressing maximization of a lower bound  
on the objective function $\objv$ over parameter values under stable synthesis constraints.
\begin{equation}\label{form:opt1}
\regmax{\varphi_M}{\objv,\theta}
\eqdef\mathop{\max\limits_{p}} \{z \mid\eta(p) \wedge
    \forall p'~
    \forall x y~[
    \theta(p,p') \implies
    (\varphi_M(p',x,y)  \implies 
     \varphi_{\mathit{cond}}^{\geq}(p',x,y,z))
    ]\}
\end{equation}
where
\[
\varphi_{\mathit{cond}}^{\geq}(p',x,y,z)
\eqdef \alpha(p',x) \implies (\beta(p',x,y) \wedge \objv(p',x,y) \geq z).
\]
The stable synthesis constraints are part of a GEAR formula and include usual $\eta, \alpha, \beta$ constraints together with the stability constraints $\theta$.
Equivalently, stable optimized synthesis can be stated as the \emph{max-min} optimization problem, Formula~\eqref{form:opt2}
\begin{equation}\label{form:opt2}
\regmax{\varphi_M}{\objv,\theta}
\eqdef
\mathop{\max\limits_{p}} \mathop{\min\limits_{x, p'}}
\{ z \mid \eta(p) \wedge
    \forall y~[
     \theta(p,p') \implies
     (\varphi_M(p',x,y)  \implies  \varphi_{\mathit{cond}}^{\leq}(p',x,y,z))]\}
\end{equation}
%
where \[
\varphi_{\mathit{cond}}^{\leq}(p',x,y,z)
\eqdef\alpha(p',x) \implies (\beta(p',x,y) \wedge \objv(p',x,y) \leq z)\text.\]
In Formula~\eqref{form:opt2} the minimization predicate in the stability region corresponds to the universally quantified $p'$ ranging over this region in~\eqref{form:opt1}.
An advantage of this formulation is that this formula can be adapted to define other aggregation functions over the objective's values on stability region.
For example, that way one can represent the \emph{max-mean} optimization problem, where one wants to maximize the mean value of the function in the stability region rather one the min value (which is maximizing the worst-case value of $f$ in stability region).
Likewise, Formula~\eqref{form:opt2} can be adapted to other interesting statistical properties of distribution of values of $f$ in the stability region.

We explicitly incorporate assertions in stable optimized synthesis by defining $\beta(p',x,y)\eqdef \beta'(p',x,y) \wedge \assert(p',x,y)$ in $\varphi_{\mathit{cond}}^{\geq}(p',x,y,z)$ of \Cref{form:opt2}, where  $\assert(p',x,y)$ are assertions required to be valid in the entire stability region around the selected configuration of knobs $p$:

\begin{equation}\label{form:optsyn}
\regmax{\varphi_M}{\objv,\theta}
\eqdef
\mathop{\max\limits_{p}} \mathop{\min\limits_{x, p'}}
\{ z \mid \eta(p) \wedge
    \forall y~[
     \theta(p,p') \implies
     (\varphi_M(p',x,y)  \implies  \varphi_{\mathit{cond}}^{\leq}(p',x,y,z))]\}
\end{equation}
where \[
\varphi_{\mathit{cond}}^{\leq}(p',x,y,z)
\eqdef\alpha(p',x) \implies ( \beta(p',x,y) \wedge \assert(p',x,y) \wedge \objv(p',x,y) \leq z)\text.\]

The notion of $\varepsilon$-solutions for these problems carries over from the one given above for Formula~\eqref{form:opt_p}.

SMLP implements stable optimized synthesis based on the GearOPT$_\delta$ and GearOPT$_\delta$-BO 
algorithms~\cite{DBLP:conf/fmcad/BrausseKK20,DBLP:conf/ijcai/BrausseKK22}, which are shown to be complete and terminating 
for this problem under mild conditions. 
These algorithms were further extended in SMLP to Pareto point computations to handle multiple objectives simultaneously.

SMLP invocation in $\mode{optsyn}$ mode is similar to running SMLP in $\mode{optimize}$ mode, discussed in \Cref{sec:example}, with the only difference that the mode is specified as $\optionval{-mode}{optsyn}$, and
the specification file or the command line for the  $\option{optsyn}$ mode should contain specification for
both assertions and objectives (while in  $\option{optimize}$ mode assertion specification is not required.)
Similarly to the $\option{optimize}$ mode, optimization progress and final results are reported in files with suffix
$\suffix{*\_optimization\_progress.\{json|csv\}}$ and  $\suffix{*\_optimization\_results.\{json|csv\}}$.

An example command for mode $\mode{optsyn}$ is given in \Cref{fig:command:optsyn}
\begin{figure}[hb!] 
\begin{verbatim}
../../src/run_smlp.py -data "../data/smlp_toy_basic" -out_dir ./ -pref Test125 \
-mode optsyn -pareto t -model system -resp y1,y2 -feat p1,p2 -save_model f \
-use_model f -mrmr_pred 0 -model_per_response t -epsilon 0.00000001 -plots f -seed 10 \
-log_time f -spec ../specs/smlp_toy_system_stable_constant_synth_feasible.spec
\end{verbatim}
\caption{Example of SMLP's command in mode $\mode{optsyn}$.}
\label{fig:command:optsyn}
\end{figure}

Figure~\ref{fig:optsyn:result} displays an example results file in $\mode{optsyn}$ mode.
\begin{figure}[tp]
\small
\begin{verbatim}
{
        "objv1": {
                "value_in_config": 0.0,
                "threshold_scaled": -0.022943015285783935,
                "threshold": 0.0,
                "max_in_data": 10.7007,
                "min_in_data": 0.24
        },
        "objv2": {
                "value_in_config": 0.0,
                "threshold_scaled": -0.010615194948015235,
                "threshold": 0.0,
                "max_in_data": 102.36396627,
                "min_in_data": 1.0752000000000002
        },
        "y2": {
                "value_in_config": 0.0,
                "value_in_system": 0
        },
        "p1": {
                "value_in_config": 0.0
        },
        "y1": {
                "value_in_config": 0.0,
                "value_in_system": 0
        },
        "p2": {
                "value_in_config": 0.0
        },
        "objv2_scaled": {
                "value_in_config": -0.010615194948015235
        },
        "threshold_lo_scaled": {
                "value_in_config": -0.010615194948015235
        },
        "threshold_lo": {
                "value_in_config": -0.010615194948015235
        },
        "threshold_up_scaled": {
                "value_in_config": -0.006959520828193561
        },
        "threshold_up": {
                "value_in_config": -0.006959520828193561
        },
        "max_in_data": {
                "value_in_config": 1.0
        },
        "min_in_data": {
                "value_in_config": 0.0
        },
        "smlp_execution": "completed",
        "interface_consistent": "true",
        "model_consistent": "true",
        "synthesis_feasible": "true"
}
\end{verbatim}
\caption{SMLP result in mode $\mode{optsyn}$.}
\label{fig:optsyn:result}
\end{figure}

\section{Design of experiments}\label{sec:doe}

Most DOE methods are based on understanding multivariate distribution of legal value combinations
of inputs and knobs in order to sample the system.
When the number of system inputs and/or knobs is large (say hundreds or more), the DOE may not  generate  
a high-quality coverage of the system’s behavior to enable training models with high accuracy.
Model training process itself becomes less manageable when number of input variables grows,
and models are not explainable and thus cannot be trusted.
One way to curb this problem is to select a subset of input features for DOE and for model training.
The problem of combining feature selection with DOE generation and model training
is an important research topic of practical interest, and SMLP supports multiple practically proven ways
to select subsets of features and feature combinations as inputs to DOE and training, including the \emph{MRMR}
feature selection algorithm~\cite{DBLP:journals/jbcb/DingP05}, and a \emph{Subgroup Discovery (SD)}
algorithm~\cite{DBLP:books/mit/fayyadPSU96/Klosgen96,DBLP:conf/pkdd/Wrobel97,DBLP:journals/widm/Atzmueller15}.
The MRMR algorithm selects a subset of features according to the principle of \emph{maximum relevance and 
minimum redundancy}. It is widely used for the purpose of selecting a subset of features for building accurate models,
and is therefore useful for selecting a subset of features to be used in DOE; it is a default choice in SMLP for that usage.
The SD algorithm selects regions in the input space relevant to the response, using heuristic statistical methods,
and such regions can be prioritized for sampling in DOE algorithms.

In the context of DOE, \emph{experiments} are lists of $\specval{(feature,value)}$ (also called  $\specval{(factor,level)}$) pairs
$$[(\specval{feature_1}, \specval{value_1}),\ldots,(\specval{feature_n}, \specval{value_n})],$$
\noindent and they are rows of the matrix of experiments returned by the supported DOE algorithms.
SMLP options that are required to invoke any of the supported DOE heuristics are:
\begin{itemize}
\item $\speckey{doe\_factor\_level\_ranges}$ A dictionary of levels per feature for building experiments for all supported DOE algorithms. 
The features are integer features (thus the values are integers). 
The keys in that dictionary are names of  features and the associated values are lists $\specval{[val_1, .., val_k]}$
from which value for that feature are selected to build an experiment. We refer to these lists of values as the \emph{grids} associated to 
each feature. DOE algorithms that work with two levels only treat these levels as the min and max of the grid range of a numeric variable.

Example: $\{\speckey{Pressure}:[50,60,70], \speckey{Temperature}:[290, 320, 350], \speckey{Flow rate}:[0.9,1.0]\}$. 

\item $\speckey{doe\_algo}$ Allows to specify the folloving DOE algorithms supported in SMLP: 
\begin{itemize}
\item $\speckey{full\_factorial}$ Builds a full factorial design dataframe from a dictionary of feature value grids,  $\specval{doe\_factor\_level\_ranges}$.
Here the attribute \emph{full} means that all combinations of $\specval{(feature,value)}$ pairs are used to build experiments, which is not feasible
with a large number of features, and with possibly more than two values (levels) in the corresponding grid of values.
\item $\speckey{fractional\_factorial}$ Builds a $2$-level fractional factorial design dataframe from a dictionary $\specval{doe\_factor\_level\_ranges}$ of 
feature value grids and given \emph{resolution}. Here the attribute \emph{$2$-level} means that every feature is represented with (or ranges over) two values only, 
and the attribute \emph{factorial} means that, unlike in $\speckey{full\_factorial}$ design, only a subset of all possible  $\specval{(feature,value)}$ pairs
are used to build experiments. A resolution is a way to specify how to group $\specval{(feature,value)}$ pairs to specify the subset of experiments to build.
\item $\speckey{plackett\_burman}$ Builds a Plackett-Burman design dataframe from a dictionary of feature value grids. Only min and max values of the range are required.
\item $\speckey{sukharev\_grid}$ Builds a Sukharev-grid hypercube design dataframe from a dictionary of feature value grids.
\item $\speckey{box\_behnken}$ Builds a Box-Behnken design dataframe from a dictionary of feature value grids.
\item $\speckey{box\_wilson}$ Builds a Box-Wilson central-composite design dataframe from a dictionary of feature value grids.
\item $\speckey{latin\_hypercube}$ Builds simple Latin Hypercube from a dictionary of feature value grids.
Latin Hypercube  sampling selects required number of sampling points so that no two samples use the same grid value, for any
of the features (so selection is history dependent). 
 Latin Hypercube sampling forces the samples drawn to correspond more closely with the input distribution, and it converges faster
than the Monte Carlo sampling which also uses random sampling from feature value distributions.
\item $\speckey{latin\_hypercube\_sf}$ Builds a space-filling Latin Hypercube design dataframe from a dictionary of feature value grids.
The attribute \emph{space filling} indicates that sampling is done after dividing feature ranges into a predefined number of
equal intervals,  which are then randomly sampled. 
\item $\speckey{random\_k\_means}$ Builds designs with $\specval{random\_k-means\_ clusters}$ from a dictionary of feature value grids.
\item $\speckey{maximin\_reconstruction}$ Builds maximin reconstruction matrix from a dictionary of feature value grids. 
\item $\speckey{halton\_sequence}$ Builds Halton matrix based design from a dictionary of  feature value grids.
\item $\speckey{uniform\_random\_matrix}$ Builds uniform random design matrix from a dictionary of feature value grids.
\end{itemize}
\item $\speckey{doe\_num\_samples}$ Number of samples (experiments) to generate.
\end{itemize}
\ZK{Add description of  $\speckey{doe\_design\_resolution}$? depends whether it is common for all or most DOE options.}

\ZK{discuss MRMR command line options? }

SMLP $\option{--help}$ provides detailed information on all DOE options. 
SMLP uses the pyDOE package \url{https://pythonhosted.org/pyDOE/} to implement DOE, and any missing details 
on DOE options and further references can be found there.

An example command for mode $\mode{doe}$ is given in \Cref{fig:command:doe}
\begin{figure}[hb!] 
\begin{verbatim}
../../src/run_smlp.py -doe_spec.csv ../grids/doe_four_levels_real -out_dir ./ \
-pref Test34 -mode doe -doe_algo full_factorial -log_time f
\end{verbatim}
\caption{Example of SMLP's command in mode $\mode{doe}$.}
\label{fig:command:doe}
\end{figure}

The generated data will be in file 
$\suffix{Test34\_doe\_four\_levels\_real\_doe.csv}$,
the top rows of which are displayed in Figure~\ref{fig:doe:result}.
\begin{figure}[tp]
\small
\centering
\begin{verbatim}
                    a,b,c                   a,b,c                 
                    5.2,-1,0.1              2.3,-1.0,0.1
                    7.1,1,1                 3.6,-1.0,0.1
                    3.6,0,2                 5.2,-1.0,0.1
                    2.3,,4.3                7.1,-1.0,0.1
                                            2.3,0.0,0.1
                                            3.6,0.0,0.1
                                            5.2,0.0,0.1
                                            7.1,0.0,0.1
                                            2.3,1.0,0.1
                    \end{verbatim}
\caption{SMLP doe spec (on the left) and part of generated data (on the right), in mode $\mode{doe}$.}
\label{fig:doe:result}
\end{figure}

\section{Root cause analysis}\label{sec:rca}

We view the problem of root cause analysis as dual to the stable optimized synthesis problem: while during optimization with 
stability we are searching for regions in the input space (or in other words, characterizing those regions) where the system response 
is good or excellent, the task of root-causing can be seen as searching for regions in the input space where the system response is 
not good (is unacceptable). Thus simply by swapping the definition of excellent vs unacceptable, we can apply SMLP to explore 
weaknesses and failing behaviors of the system.

Even if a number of witnesses (counter-examples to an assertion) are available, they represent discrete points in the input space 
and it is not immediately clear which value assignments to which variables in these witnesses are critical to explain the failures. 
Root causing capability in SMLP is currently supported through two independent approaches:
a \emph{Subgroup Discovery (SD)} algorithm that searches through the data for the input regions where there is a higher ratio 
(thus, higher probability) of failure; to be precise, SD algorithms support a variety of \emph{quality functions} which play the role of 
optimization objectives in the context of optimization. 
See \cite{DBLP:conf/fmcad/BrausseKK20,DBLP:conf/dac/KatzRZS11,DBLP:conf/ispd/Wang13,DBLP:journals/ijdsa/KhasidashviliN21,DBLP:journals/corr/abs-2207-00622}
for usage of SD and closely related techniques of \emph{Rule Learning} in validation and test, including for root-causing.
To find input regions with high probability of failure, SMLP searches for stable witnesses to failures. These capabilities, together with 
feature selection algorithms supported in SMLP, enable researchers to develop new root causing capabilities that combine formal 
methods with statistical methods for root cause analysis. 

An example command for mode $\mode{subgroups}$ is given in \Cref{fig:command:subgroups}.
The generated sungroup descriptions are reported in file with suffix $\speckey{*\_features\_ranking.csv}$.
\begin{figure}[hb!] 
\begin{verbatim}
../../src/run_smlp.py -data ../data/smlp_toy_num_resp_mult -out_dir ./ \
-pref Test30 -mode subgroups -psg_dim 3 -psg_top 10 -resp y1,y2 -feat x,p1,p2 \
-plots f -seed 10 -log_time f
\end{verbatim}
\caption{Example of SMLP's command in mode $\mode{subgroups}$.}
\label{fig:command:subgroups}
\end{figure}

\section{Model refinement loop}\label{sec:refinement}

Support in SMLP for selecting DOE vectors to sample the system and generate a training set was discussed in Subsection~\ref{sec:doe}.
Initially, when selecting sampling points for the system, it is unknown which regions in the input space are really relevant for the exploration 
task at hand. Therefore some DOE algorithms also incorporate random sampling and
sampling based on previous experience and familiarity with the design, such as sampling nominal cases and corner cases, when these are known.
For model exploration tasks supported by SMLP, it is not required to train a model that will be an accurate match to the system everywhere in 
the legal search space of inputs and knobs.
We require to train a model that is an \emph{adequate} representation of the system for the task at hand, 
meaning that the exploration task solved on the model solves this task for the system as well.
Therefore SMLP supports a \emph{targeted model refinement} loop to
enable solving the system exploration tasks by solving these tasks on the model instead.
The idea is as follows: when a stable solution to model exploration task is found, it is usually the case that there are not 
many training data points close to the stability region of that solution.
This implies that there is a high likelihood that the model does not accurately represent the system in the stability region of the solution.
Therefore the system is sampled in the stability region of the solution, and these data samples are added to the initial training data to 
retrain the model and make it more adequate in the stability region of interest.
Samples in the stability region of interest can also be assigned higher weights compared to other samples to help training to achieve 
higher accuracy in that region. More generally, higher adequacy of the model can be achieved by sampling distributions biased towards 
prioritizing the stability region during model refinement.

Note that model refinement is required only to be able to learn properties on the system by exploring these properties on the model.
As a simple scenario, let us consider that we want to check an assertion $\assert(p,x,y)$ on the system.
If a $\theta$-stable counter example $x^*$ to $\assert(p,x,y)$ for a configuration $p^*$ of knobs exists 
on the model according to Definition~\ref{def:stable:witness:validity} 
(which means that  $x^*$ is a $\theta$-stable witness for query $\query(p^*, x, y) \eqdef \neg \assert(p^*,x,y)$), 
then the system is sampled in the $\theta$--stability region of $p^*$ (possibly with $x^*$ toggled as well in a small 
region around $x^*$). If the failure of assertion $\assert(p,x,y)$ is reproduced on the system using the stable witness 
$x^*$ and a configuration in the $\theta$-stability region of $p^*$, then the model exploration goal has been 
accomplished (we found a counter-example to $\assert(p,x,y)$ on the system) and model refinement can stop.
Otherwise the system samples can be used to refine the model in this region. For that reason, the wider the 
$\theta$-stability region, the higher the chances to reproduce failure of assertion $\assert(p,x,y)$ on the system.

On the other hand, if  $\assert(p,x,y)$ does not have a $\theta$-stable counter-example, then it is still possible 
that $\assert$ is not valid on the system but it cannot be falsified on the model due to discrepancy between the system 
and model responses in some (unknown to us) input space. In this case one can strengthen $\assert(p,x,y)$ to $\assert'(p,x,y)$
(for example, assertion $\assert(p,x,y) \eqdef y \geq 3$ can be  strengthened to $\assert'(p,x,y) \eqdef y \geq 3.01$),
or one can weaken the stability condition $\theta_r$ to  $\theta_{r'}$ by shrinking the stability radii $r$ to $r'$; or do both.
If the strengthened assertion $\assert'(p,x,y)$ has a $\theta_{r'}$-stable counter-example $x'$ for a configuration $p'$ on 
the model, then $p', x'$ can be used to find a counter-example to the original assertion $\assert(p, x, y)$ on the system in the same 
way as with $p^*, x^*$ before. If failure of the original assertion $\assert$ is reproduced on the system, the model 
refinement loop stops, otherwise it can continue using other strengthened versions of the original assertion and or shrinking the  
stability radii even further. Similar reasoning is applied to other modes of design space exploration. In particular, in the 
optimization modes, one can confirm or reject on the system an optimization threshold proved on the model, and in the latter 
case the model refinement loop will be triggered by providing new data-points 
that can be used for model refinement  in this region, through re-training or incremental training of a new model. 
\KK[inline]{Might be move to the end of 1st para: Similar reasoning is applied to other modes of design space exploration. 
In particular, in the optimization modes, the region around optimal solution can be sampled and simulated by the system, 
and either confirm the model predictions or provide new data-points that can be used for model refinement in this region.}


\bibliography{bib}

\delete{

}

\delete{%
\appendix
\section{Concrete Formats}
\subsection{CSV}\label{sec:csv}
A structured line-based text format where the first line describes a
list of column headers and each line after it describes a list the
values corresponding to the respective column in the data set.
A line describes a list of elements $(e_i)_{i=1}^n$, if it corresponds
to the concatenation
$s(e_1)\literal,s(e_2)\literal,\ldots\literal,s(e_n)$
where $s(v)$ is the ASCII representation of $v$.

The list of column headers shall not contain duplicates.
The ASCII representation of ASCII strings made up of printable
characters excluding ``\literal,'' is the identity.
Floating point approximations are represented as the decimal encoding
where ``\literal.'' is the separator between integer and fractional
part. Rational numbers $\frac pq$ are represented by $s(p)\literal/s(q)$
where $s(z)$ is the decimal encoding of $|z|$ prefixed by ``\literal-''
if $z\in\mathbb Z$ is negative.

\section{Concepts and Preliminaries}

In this document, syntactical highlights are placed on
\literal{literal strings}, which are ASCII sequences used to communicate input
to and output from programs; on \cmd{commands} to be executed;
and on \api{API references} which correspond to
either concrete symbols or abstract concepts defined in the corresponding API.

We assume familiarity with JSON, \cmd{make} and CSV files. Since the CSV format
is not unambiguously defined, we give the concrete requirements in
\cref{sec:csv}.

The glossary in this section defines concepts which are used throughout this
document when describing in detail the problems solved by \SolverAbbrv and is
meant to be referred to on an per-instance basis instead of reading it as a
whole.
\begin{description}
\item[Center Threshold]
	A rational value in $[0,1]$ larger or equal to Threshold.
\item[Codomain]
	A real vector space.
\item[Data set]
	A list of points in $D\times C$ where $D$ is the domain and $C$ is the
	codomain.
\item[Domain]
	A domain $D$ is the cartesian product $\bigtimes_{i=1}^n D_i$ where
	$D_i$ is a subset of either $\mathbb Z$, $\mathbb R$ or a
	discrete finite subset of $\mathbb Q$. \ZK{In other places we use sets instead of finite subsets of Q?}
\item[Feature]
	Any column of a data set $\mathcal D$ is called a feature.
\item[Instance]
	A tuple
	$(N,\mathcal N_D,o,\mathcal N_o)$ is called an instance if
	$N$ is an NN over domain $D$, $\mathcal N_D$ is a data normalization,
	$o$ is an objective function and
	$\mathcal N_O$ is an objective normalization.
	If -- in addition to an instance $I$ -- a data set $\mathcal D$ over $D$
	is given, $(I,\mathcal D)$ is called a data instance.
\item[NN]
	Neural network as understood by the Keras API of Tensorflow in HDF5
	format.
	In particular, the \Solver only handles \api{Dense} layers with either
	\api{relu} or \api{linear} activation function.
	The input layer must be defined on domain $D$ and the output layer
	must map to the codomain $C$.
\item[Region]
	Given a domain $D=\bigtimes_{i=1}^d D_i$, a region (in $D$) is a
	product $\bigtimes_{i=1}^d C_i$ where
	each $C_i\subseteq D_i$ is a finite union of compact intervals for
	$i=1,\ldots,n$.
	For discrete $D_i$, e.g., a subset of $\mathbb Z$,
	$C_i$ is the finite union of point intervals.
	Otherwise, $D_i$ is a bounded subset of $\mathbb R$ and $C_i$
	corresponds to just one interval $[c_i\pm r_i]$ with center $c_i\in D_i$
	and rational radius $r_i>0$.
\item[Safe]
	A region $R$ is considered safe (wrt.\ a given target function $f$) for a
	constant $t\in\mathbb Q$ if $f$ satisfies $f(x)\geq t$
	for all $x\in R$.
\item[.spec]
	Specification file describing the domain, codomain and regions in the
	domain. Its format is a JSON list where the $i$-th entry corresponds to
	the $i$-th column in the CSV describing the data set.

	Given a data set and a .spec file, the components $D_i$ of the domain
	are defined as $D_i=E_i(F_i)$ where
	$E_i$ is called the embedding of $F_i$ into $D_i$ where $F_i$ is the
	$i$-th input feature.
\item[Target function]
	The target function $f_I$ of an instance
	$I=(N,\mathcal N_D,o,\mathcal N_o)$ is defined as the composition
	$\mathcal N_o\circ o\circ\mathcal N_{D,\mathrm o}\circ
	N\circ\mathcal N_{D,\mathrm i}$.
	The target function $f_{I,\mathcal D}$ of a data instance
	$(I,\mathcal D)$ is
	$x\mapsto\min(\{f_I(x)\}\cup\{\mathcal N_o(o(y)):(x,y)\in\mathcal D\})$.
\item[Threshold]
	The threshold for a region $R$ and target function $f$ is defined
	as the maximal $t\in\{t_1,\ldots,t_n\}\subset[0,1]\cap\mathbb Q$ for
	which $R$ is safe wrt.\ $f$ for $t$ if it exists, and $-\infty$ otherwise.

	The threshold for a domain $D$ and target function $f$ is defined
	as the maximal $t\in\{t_1,\ldots,t_n\}\subset[0,1]\cap\mathbb Q$ for
	which there is a region that is safe wrt.\ $f$ for $t$ if it exists, and
	$-\infty$ otherwise.
\end{description}
}%

\end{document}